\documentclass{article}

\usepackage{arxiv}
\usepackage[margin=1in]{geometry}
\usepackage{microtype}
\usepackage{graphicx}
\usepackage{subfigure}
\usepackage{booktabs} % for professional tables
\usepackage{natbib}
\usepackage[affil-it]{authblk} 
\usepackage{enumitem}

\usepackage{fontawesome}

\usepackage{amsmath,amsfonts,bm}

\def\eqref#1{equation~\ref{#1}}
\def\1{\bm{1}}

\DeclareMathAlphabet{\mathsfit}{\encodingdefault}{\sfdefault}{m}{sl}
\SetMathAlphabet{\mathsfit}{bold}{\encodingdefault}{\sfdefault}{bx}{n}

\usepackage{subfiles}
\usepackage{amsmath}
\usepackage[colorlinks,citecolor=blue,urlcolor=blue,linkcolor=blue,linktocpage=true]{hyperref}
\usepackage[capitalize]{cleveref}
\usepackage{libertine}
\usepackage{libertinust1math}
\usepackage[T1]{fontenc}
\usepackage{multirow}
\usepackage{makecell}
\usepackage[rgb,dvipsnames]{xcolor} % 支持颜色高亮
\usepackage{algorithm}
\usepackage{algpseudocode}
\usepackage{algorithmicx}

\newcommand{\datascaling}[1]{\textcolor{OliveGreen}{\textbf{#1}}}    % Data Scaling (蓝色)
\newcommand{\networkscaling}[1]{\textcolor{OrangeRed}{\textbf{#1}}}  % Network Scaling (红色)
\newcommand{\budgetscaling}[1]{\textcolor{RoyalBlue}{\textbf{#1}}} % Training Budget Scaling (绿色)

\usepackage[textsize=tiny,
 disable
]{todonotes}

\usepackage{soul}

\newcommand{\revisedelete}[1]{}

\allowdisplaybreaks

\title{Scaling DRL for Decision Making: A Survey on Data, Network, and Training Budget Strategies}

\author{
{Yi Ma}$^{1,}\thanks{Equal contributions.}$ \ \
Hongyao Tang$^{2,*}$\ \ 
{Chenjun Xiao}$^3$\ \  
{Yaodong Yang}$^4$ \ \
{Wei Wei}$^1$\ \  
{Jianye Hao}$^2$\ \  
{Jiye Liang}$^1$\ \  
}
\affil{
$^1$School of Computer and Information Technology, Shanxi University \\
$^2$College of Intelligence and Computing, Tianjin University \\
$^3$School of Data Science, The Chinese University of Hong Kong (Shenzhen) \\
$^4$Department of Computer Science and Engineering, The Chinese University of Hong Kong \\
 {mayi@sxu.edu.cn},\ 
 {tanghongyao@tju.edu.cn},\ 
 {chenjunx@cuhk.edu.cn},
 {yangyaodong@link.cuhk.edu.hk},\\
 {weiwei@sxu.edu.cn},\
 {jianye.hao@tju.edu.cn},\
 {ljy@sxu.edu.cn}
 
}

\begin{document}

\maketitle

\begin{abstract}

In recent years, the expansion of neural network models and training data has driven remarkable progress in deep learning, particularly in computer vision and natural language processing. This advancement is underpinned by the concept of Scaling Laws, which demonstrates that increasing model parameters and training data enhances learning performance. While these fields have witnessed transformative breakthroughs, such as the development of large language models like GPT-4 and advanced vision models like Midjourney, the application of scaling laws in deep reinforcement learning (DRL) remains relatively unexplored. Despite its potential to significantly improve model performance, stability, and generalization, the integration of scaling laws into DRL for decision making has not been fully realized. This review addresses this gap by systematically analyzing scaling strategies in three key dimensions: data, network, and training budget. In data scaling, we explore methods to optimize data efficiency through parallel sampling and data generation, examining the relationship between data volume and learning outcomes. For network scaling, we investigate architectural enhancements, including width and depth expansions, ensemble and MoE methods, and agent number scaling techniques, which collectively enhance model expressivity while posing unique computational challenges. Lastly, in training budget scaling, we evaluate the impact of distributed training, high replay ratios, large batch sizes, and auxiliary training on training efficiency and convergence. By synthesizing these strategies, this review not only highlights their synergistic roles in advancing DRL for decision making but also provides a roadmap for future research. We emphasize the importance of balancing scalability with computational efficiency and outline promising directions for leveraging scaling laws to unlock the full potential of DRL in various complicated tasks such as robot control, autonomous driving and LLM training.

\end{abstract}

\section{Introduction}

The rapid advancement of deep learning over the past decade has been propelled by the systematic application of Scaling Laws~\citep{kaplan2020scaling,henighan2020scaling,hoffmann2022training,Marafioti25smol} that establish predictable relationships between model performance and computational resources. By scaling model parameters, training data, and computational budgets, breakthroughs in domains such as computer vision and natural language processing (NLP) have redefined state-of-the-art capabilities. Transformative models like GPT-4~\citep{OpenAI23gpt4} and Midjourney\footnote{\url{https://www.midjourney.com}} exemplify the power of scaling, achieving unprecedented generalization and creativity through massive architectures and expansive datasets. Yet, while these fields have embraced scaling as a cornerstone of progress, its integration into deep reinforcement learning (DRL) remains nascent, leaving a critical gap in the pursuit of robust, generalizable, and efficient intelligent systems.

\begin{figure}[!htbp]
\centering
\includegraphics[scale=0.6]{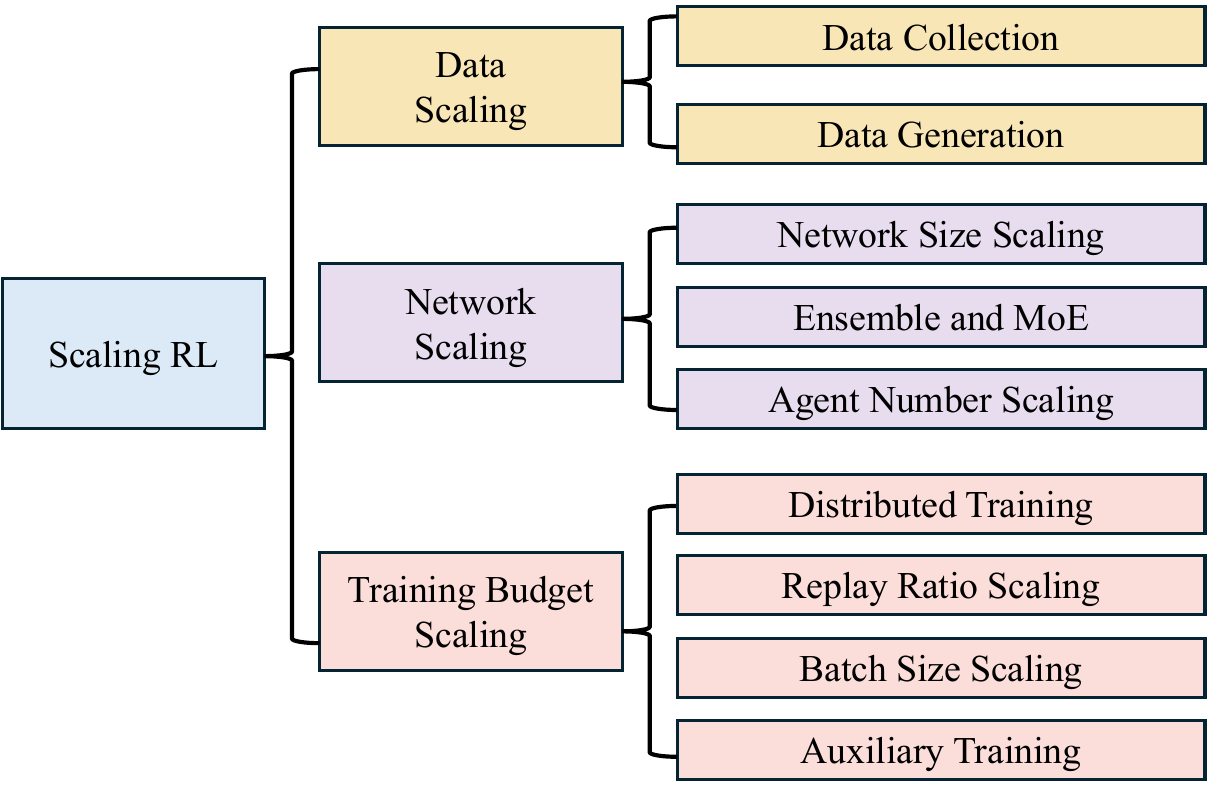}
\caption{Illustration of our taxonomy of the current literature on Scaling DRL.}
\label{fig:taxonomy}
% \vspace{-0.5cm}
\end{figure}

DRL’s strengths lie in its ability to learn complex behaviors through trial-and-error interactions, leveraging deep neural networks to approximate policies and value functions in high-dimensional state-action spaces. Techniques like Q-learning \citep{mnih2015human}, policy gradients \citep{schulman2017proximal}, and actor-critic methods \citep{haarnoja2018soft} have achieved remarkable success in domains ranging from robotics to game playing. While recent works \citep{hilton2023scaling, rybkin2025value}  have demonstrated that scaling neural networks and training data can enhance RL's decision making performance in specific settings, a systematic understanding of how scaling laws govern learning dynamics, stability, and generalization remains elusive. This gap persists despite evidence that strategic scaling could address longstanding RL challenges, such as sample inefficiency, reward sparsity, and catastrophic forgetting. Therefore, this review synthesizes emerging insights into scaling strategies for DRL, focusing on \textbf{three pivotal dimensions}: data, network architectures, and training budgets as shown in Figure \ref{fig:taxonomy}. 

Data scaling enhances reinforcement learning through parallelized collection and synthetic generation, addressing sample inefficiency and limited exploration. Distributed frameworks like Ape-X \citep{horgan2018distributed} decouple actors and learners to achieve near-linear throughput scaling, while SAPG \citep{singla2024sapg} partitions environments for diversified exploration. Synthetic data methods, such as SYNTHER \citep{lu2023synthetic} and HIPODE \citep{lian2023hipode}, use diffusion models and value-guided filtering to generate high-fidelity transitions, augmenting offline datasets and reducing reliance on real interactions. Together, these approaches expand data volume and coverage, enabling efficient training in complex tasks with sparse rewards.

Network scaling unlocks performance gains through architectural expansion, ensemble diversification  and agent number increasing. Wider residual MLPs with different normalizations (SimBa \citep{lee2024simba}) and transformer-based architectures (Gato \citep{reedgeneralist}) improve expressivity and cross-task generalization. Ensemble methods like REDQ \citep{chenrandomized} mitigate overestimation via randomized subsampling, while Mixture-of-Experts (MoE \citep{obando2024mixtures}) dynamically scales network capacity. Agent number scaling further  enhance exploration. The representative method is Evolutionary RL \citep{khadka2018evolution}, which introduces population-based mutation, and  achieves  higher sample efficiency in sparse-reward domains. These approaches collectively demonstrate that scaled networks not only enhance individual model performance but also unlock emergent behaviors through modularity and diversity.

Training budget scaling optimizes resource allocation through distributed parallelism, data reuse, batch scaling and auxiliary learning. IMPALA \citep{espeholt2018impala} leverages decoupled actor-learners for high-throughput training, while BRO \citep{naumanbigger} combines layer normalization and large critics to tolerate aggressive update-to-data (UTD) ratios. Batch scaling methods like LaBER \citep{lahire2022large} reduce gradient variance via importance-weighted sampling, and auxiliary objectives (SPR \citep{SchwarzerAGHCB21SPR}) inject task-agnostic structure to accelerate convergence. These innovations prioritize compute efficiency by balancing sample
efficiency, algorithmic robustness, and hardware utilization across diverse learning scenarios.

\begin{table*}[!htbp]
\centering
\caption{Summarization of scaling axes of representative works.}
\label{tab:summary}

\scalebox{0.95}{
\begin{tabular}{c|ccc}
\toprule
\textbf{Method} & \textbf{Data Scaling} & \textbf{Network Scaling} & \textbf{Training Budget Scaling} \\ 
\midrule
\textbf{Ape-X} \citep{horgan2018distributed} & Data Collection & - & -  \\
\textbf{OpenAI Five} \citep{berner2019dota} & Data Collection & - & Replay Ratio \\
\textbf{AlphaStar} \citep{vinyals2019grandmaster} & Data Collection & & Batch Size, Auxiliary Training  \\
\textbf{PQN} \citep{gallici2024simplifying} & Data Collection & - &  - \\
\textbf{FastTD3} \citep{seo2025fasttd3} & Data Collection & - & Batch Size \\
\textbf{SAPG} \citep{singla2024sapg} & Data Collection & - & -  \\
\textbf{ExORL} \citep{yarats2022don} & Data Collection & - & -  \\
\textbf{Scaled QL}\citep{kumaroffline} & Data Collection & Network Size & Auxiliary Training   \\
\textbf{SYNTHER} \citep{lu2023synthetic} & Data Generation & - & Replay Ratio  \\
\textbf{PGR} \citep{wang2024prioritized} & Data Generation & - & Replay Ratio  \\
\textbf{OFENet} \citep{ota2020can} & - & Network Size & -  \\
\textbf{BRO} \citep{naumanbigger} & - & Network Size & Replay Ratio   \\
\textbf{BRC} \citep{nauman2025bigger} & Data Collection & Network Size & Replay Ratio   \\
\textbf{Simba} \citep{lee2024simba} & - & Network Size & Replay Ratio   \\
\textbf{Simbav2} \citep{lee2025hyperspherical} & - & Network Size & Replay Ratio   \\
\textbf{BBF} \citep{schwarzer2023bigger} & Data Collection & Network Size & Replay Ratio   \\
\textbf{Scaling CRL} \citep{wang20251000} & - & Network Size & Replay Ratio, Batch Size   \\
\textbf{DT-VINs} \citep{wang2024scaling} & - & Network Size & -   \\
\textbf{GTrXL} \citep{parisotto2020stabilizing} & Data Collection & Network Size &  - \\
\textbf{PAC} \citep{springenberg2024offline} & Data Collection & Network Size & -  \\
\textbf{HL-Gauss} \citep{farebrother2024stop} & - & Network Size & -  \\
\textbf{CHAIN} \citep{tang2024improving} & - & Network Size & -  \\
\textbf{Bootstrapped DQN} \citep{osband2016deep} & - & Ensemble & -  \\
\textbf{SUNRISE} \citep{lee2021sunrise} & - & Ensemble & -  \\
% \textbf{Maxmin Q-learning} \citep{lanmaxmin} & - & Ensemble & -  \\
\textbf{EBQL} \citep{peer2021ensemble} & - & Ensemble & -  \\
\textbf{TQC} \citep{kuznetsov2020controlling} & - & Ensemble & -  \\
\textbf{REDQ} \citep{chenrandomized} &  -& Ensemble & Replay Ratio   \\
\textbf{DroQ} \citep{hiraokadropout} & - & Ensemble & Replay Ratio   \\
\textbf{MED-RL} \citep{sheikh2022maximizing}  & - & Ensemble & Replay Ratio   \\
\textbf{EDAC} \citep{an2021uncertainty}  & - & Ensemble & -   \\
\textbf{RLPD} \citep{ball2023efficient}  & - & Ensemble & Replay Ratio    \\
\textbf{MoE} \citep{obando2024mixtures} & -  & Ensemble & Replay Ratio   \\
\textbf{POLTER} \citep{schubertpolter} & - & Ensemble & -  \\
% \textbf{SEERL} \citep{saphal2021seerl} & - & Ensemble & -  \\
\textbf{EPPO} \citep{yang2022towards} & - & Ensemble & -  \\
\textbf{ERL} \citep{khadka2018evolution} & Data Collection & Agent Number &  - \\
\textbf{CERL} \citep{khadka2019collaborative} & Data Collection & Agent Number &  - \\
% \textbf{CSPC} \citep{DBLP:conf/nips/ZhengW0L0Z20} & Data Collection & ERL &  - \\
\textbf{ERL-Re$^2$} \citep{Re2} & Data Collection & Agent Number &  - \\
\textbf{EvoRainbow} \citep{Li0TFH24EvoRainbow} & Data Collection & Agent Number &  - \\
\textbf{IMPALA} \citep{espeholt2018impala} & Data Collection & Network Size & Distributed Training  \\
\textbf{QT-OPT} \citep{kalashnikov2018scalable} & Data Collection & - & Distributed Training  \\
\textbf{SR-SAC/SR-SPR} \citep{dsample} & - & - & Replay Ratio  \\
\textbf{MAD-TD} \citep{voelcker2024mad} & Data Generation & - & Replay Ratio  \\
\textbf{SMR} \citep{lyu2024off} & - & - & Replay Ratio  \\
\textbf{CrossQ+WN} \citep{palenicek2025scaling} & - & Network Size & Replay Ratio  \\
\textbf{MARR} \citep{yang_sample-efficient_2024} & - & - & Replay Ratio  \\
\textbf{LaBER} \citep{lahire2022large} & - & - & Batch Size  \\
\textbf{CURL} \citep{LaskinSA20CURL} & - & - & Auxiliary Training  \\
\textbf{UNREAL} \citep{JaderbergMCSLSK17UNREAL} & - & - & Auxiliary Training  \\
\textbf{DBC} \citep{zhou2022learning} & - & - & Auxiliary Training  \\
\bottomrule
\end{tabular}
}
\end{table*}

We summarize different scaling axes of representative works in Table \ref{tab:summary}. By charting these interconnected strategies, this review advances four key contributions: 1) We give a comprehensive survey on scaling in DRL for decision making with a novel taxonomy for the first time. 2) We analyze the strengths and weaknesses of representative works on scaling DRL along with their abilities in addressing different challenges. 3) We highlight the insights of scaling RL and summarize its successful application in LLM training with critical assessment of unresolved challenges. 4) We present some practical guidelines for resource-aware DRL system design, balancing computational efficiency with performance. By synthesizing empirical findings  principles, this work establishes a foundation for developing robust scaling laws tailored to DRL’s unique constraints—a crucial step toward generalizable, efficient, and scalable reinforcement learning systems.

\section{Preliminaries}

\subsection{Markov Decision Process}

A Markov Decision Process (MDP) is a mathematical framework for modeling decision-making problems where outcomes are partly random and partly under the control of a decision-maker. An MDP is defined by a tuple $(\mathcal{S}, \mathcal{A}, \mathcal{P}, \mathcal{R}, \gamma)$, where:
\begin{itemize}
    \item $\mathcal{S}$ is the state space.
    \item $\mathcal{A}$ is the action space.
    \item $\mathcal{P}$: $\mathcal{S} \times \mathcal{A} \times \mathcal{S} \rightarrow [0, 1]$ is the transition probability function, where $\mathcal{P}(s' | s, a)$ denotes the probability of transitioning to state $s'$ after taking action $a$ in state $s$.
    \item $\mathcal{R}: \mathcal{S} \times \mathcal{A} \rightarrow \mathbb{R}$ is the reward function, which provides immediate feedback to the agent.
    \item $\gamma \in [0, 1)$ is the discount factor that determines the present value of future rewards.
\end{itemize}

The goal in an MDP is to find a policy $\pi: \mathcal{S} \rightarrow \mathcal{A}$ that maximizes the expected cumulative discounted reward:
\begin{align}
    J(\pi) = \mathbb{E}_{\tau \sim \pi} \left[ \sum_{t=0}^{\infty} \gamma^t r_t \right],
\end{align}
where $\tau = (s_0, a_0, r_0, s_1, a_1, r_1, \ldots)$ is a trajectory generated by policy $\pi$.

\subsection{Reinforcement Learning}
Reinforcement Learning (RL)~\citep{Sutton1988ReinforcementLA} involves an agent learning to make decisions by interacting with an environment to maximize cumulative rewards. Traditional RL methods face challenges in high-dimensional state-action spaces and require significant feature engineering. Deep Reinforcement Learning (DRL) addresses these limitations by leveraging deep neural networks to automatically learn representations and policies. Below are three primary approaches of DRL:

\paragraph{Value-Based Methods.}
Value-based methods focus on learning a value function that estimates the expected return for each state-action pair. Q-learning is a prominent example, which updates the Q-value using:
\begin{align}
Q(s, a) \leftarrow Q(s, a) + \alpha \left[ r + \gamma \max_{a'} Q(s', a') - Q(s, a) \right],
\end{align}
where $\alpha$ is the learning rate. Among the value-based methods, the representative ones are DQN\citep{mnih2015human}, DDQN\citep{van2016deep}, Rainbow\citep{hessel2018rainbow}, etc.

\paragraph{Policy-Based Methods.}
Policy-based methods directly optimize the policy $\pi(a|s)$ to maximize the expected reward. The policy gradient theorem states that the gradient of the expected reward can be estimated as:
\begin{align}
\nabla_{\theta} J(\theta) = \mathbb{E}_{s \sim \rho^{\pi}, a \sim \pi_{\theta}} \left[ \nabla_{\theta} \log \pi_{\theta}(a|s) Q^{\pi_{\theta}}(s, a) \right],
\end{align}
where $\rho^{\pi}$ is the state distribution induced by policy $\pi$. Representative methods include TRPO\citep{schulman2015trust}, PPO\citep{schulman2017proximal}, and so on.

\paragraph{Actor-Critic Methods.}
Actor-critic methods combine value-based and policy-based approaches. The actor updates the policy based on the critic's evaluation of the actions. The critic estimates the value function, often using a Q-network, and the actor updates its parameters using the policy gradient:
\begin{align}
\nabla_{\theta} J(\theta) \approx \mathbb{E}_{s \sim \rho^{\pi}, a \sim \pi_{\theta}} \left[ \nabla_{\theta} \log \pi_{\theta}(a|s) \hat{Q}(s, a) \right],
\end{align}
where $\hat{Q}(s, a)$ is the estimated Q-value from the critic. DDPG \citep{lillicrap2015continuous}, TD3\citep{fujimoto2018addressing}, and SAC \citep{haarnoja2018soft} are the most widely used actor-critic algorithms.

\subsection{Scaling Laws in Deep Learning} 
Scaling laws~\citep{kaplan2020scaling,henighan2020scaling,hoffmann2022training,Marafioti25smol} in deep learning establish quantitative relationships between the performance of artificial neural networks and three fundamental resources: model size ($N$), dataset size ($D$), and computational budget ($C$). These empirical principles reveal a \textbf{power-law dependency}, where performance metrics (e.g., test loss, accuracy) improve predictably as resources scale, independent of architectural refinements. The systematic nature of these relationships has profound implications for model design and resource allocation in modern AI systems.  

The scaling behavior can be formalized through a multi-resource power-law equation:  
\begin{align}
P(N, D, C) = \underbrace{\alpha N^{-\beta}}_{\text{Model Scaling}} + \underbrace{\gamma D^{-\delta}}_{\text{Data Scaling}} + \underbrace{\epsilon C^{-\zeta}}_{\text{Compute Scaling}} + L_0
\end{align}  
where $P$ indicates the mode performance, $N$ is the number of trainable parameters, $D$ is the number of training samples, $C$ is the Floating-point operations (FLOPs) during training, $L_0$ is the irreducible loss floor determined by task complexity, $\alpha, \gamma, \epsilon$ are scaling coefficients, $\beta, \delta, \zeta \in (0,1)$ are scaling exponents governing diminishing returns. This separable formulation implies that each resource contributes independently to performance improvement, with the dominant term shifting based on relative scaling rates.

Several researches verify the scaling law from empirical validation. Model parameter scaling ($N$-scaling)  has been thoroughly investigated.
The relationship between performance and model parameter, i.e., $P \propto N^{-\beta}$ has been consistently observed across architectures. For language models, \cite{kaplan2020scaling} show that $\beta \approx 0.076$ for autoregressive transformers. For vision models, \cite{henighan2020scaling} show that $\beta \approx 0.089$ in CNNs. Some critical insight such as doubling model size reduces loss by a constant multiplicative factor, regardless of initial $N$ have also been summarized. As for data scaling ($D$-scaling), the scaling law $L \propto D^{-\delta}$ exhibits domain-specific variation. For example, \cite{hoffmann2022training} demonstrates that $\delta \approx 0.34$ for LLMs. 

% \subsection{Pioneer Exploration in Scaling RL}

In DRL, very few works systematically analyzed the scaling phenomenon. \cite{hilton2023scaling} introduced the concept of intrinsic performance, defined as the minimum computational resources required to achieve a given RL return, to extend neural scaling laws to single-agent RL. Through empirical analysis across diverse environments (Procgen, Dota 2, MNIST), the authors demonstrate that intrinsic performance follows power-law scaling with model size and environment interactions,  akin to generative models. The study further identifies that task horizon length primarily affects scaling coefficients rather than exponents, providing a framework to analyze computational trade-offs and sample efficiency in RL systems. \cite{rybkin2025value} further demonstrated that value-based DRL methods exhibit predictable scaling properties through systematic analysis of hyperparameter interactions and resource allocation. By establishing empirical scaling laws that link UTD ratios to Pareto-optimal tradeoffs between data efficiency and computational requirements, the authors show how to extrapolate performance from small-scale experiments to optimize large-scale training configurations. Their framework, validated across SAC, BRO, and PQL algorithms on diverse benchmarks (DMC\citep{tassa2018deepmind}, OpenAI Gym\citep{brockman2016openai}, IsaacGym\citep{makoviychuk2021isaac}), challenges prevailing assumptions about value-based RL's pathological scaling by revealing power-law relationships between hyperparameters, budget allocation, and task performance. However, these works focus on  specifically chosen environments or models, limiting their wide adaptability.

% In addition,  Notably, data quality governs effective scaling:  
% \begin{align}
% \delta_{\text{effective}} = \delta \cdot \mathbb{E}[I(x;y)]  
% \end{align}  
% where $I(x;y)$ is the mutual information between inputs $x$ and labels $y$.  
% Lastly, researches on compute scaling ($C$-scaling) shows that the compute-performance relationship $L \propto C^{-\zeta}$ depends on training parallelism:
% \begin{align}
% \zeta = \frac{1}{1 + \frac{\log(\text{Parallelizability})}{\log(\text{Batch Size Ratio})}}
% \end{align}  
% Empirical studies suggest $\zeta \approx 0.05 - 0.1$ for standard transformer training (Sohl-Dickstein et al., 2020).  

\section{Scaling DRL and its challenges}

To help understand the scaling concept in DRL, we first present different scaling dimensions in DRL training pipelines. The modified RL algorithm in Algorithm~\ref{algo:example} systematically incorporates three scaling paradigms --- \datascaling{data scaling}, \networkscaling{network scaling}, and \budgetscaling{training budget scaling} --- across distinct phases of training. This holistic approach addresses key bottlenecks in sample efficiency, model capacity, and computational resource utilization, enabling more performant and adaptable RL systems.  

\begin{algorithm}[!htbp]
\label{algo:example}
\caption{Example of RL Algorithm with Different Scaling Dimensions}
\begin{algorithmic}[1]
  \State \textbf{Initialize:}
    \State \quad Policy Network $\pi_{\theta}$ \networkscaling{(Network Size Scaling / Ensemble / MoE )} 
    \State \quad Value Network $V_{\phi}$ \networkscaling{(Network Size Scaling / Ensemble / MoE)} 
    \State \quad Replay Buffer $\mathcal{D}$ 
    \State \quad Agent Population $\mathcal{A}$ 
    \networkscaling{(Agent Number Scaling)}
    \State \quad Replay Ratio $K$ 
    
\For{each\_agent from $\mathcal{A}$ in an environment}:
  \For{episode $= 1$ \textbf{to} $M$}
    \State \textbf{// Environment Interaction}
    \State Reset environment, observe initial state $s_0$
    \For{$t = 1$ \textbf{to} $T$} \datascaling{(Data Collection)}
      \State Sample action $a_t \sim \pi_{\theta}(a|s_t)$
      \State Execute $a_t$, receive reward $r_t$, next state $s_{t+1}$
      \State Generate transition using generative models \datascaling{(Data Generation)} 
      \State Store transition $(s_t, a_t, r_t, s_{t+1})$ in $\mathcal{D}$ 
    \EndFor

    \State \textbf{// Model Update}
    \For{$k = 1$ \textbf{to} $K$} \budgetscaling{(Replay Ratio Scaling)}
      \State Sample batch $\mathcal{B} \sim \mathcal{D}$ \budgetscaling{(Batch Size Scaling)}
      \State Compute target $y_i = r_i + \gamma V_{\phi}(s_{i+1})$
      \State Update $V_{\phi}$: $\min_{\phi} \sum_i (V_{\phi}(s_i) - y_i)^2$ \budgetscaling{(Distributed Training)}
      \State Update $\pi_{\theta}$: $\max_{\theta} \sum_i \log \pi_{\theta}(a_i|s_i) \cdot A(s_i, a_i)$ \budgetscaling{(Distributed Training)}
      \State Apply auxiliary training loss \budgetscaling{(Auxiliary Training)}
    \EndFor
  \EndFor
  
   \State Evolve new agents into $\mathcal{A}$ via evolutionary strategy \networkscaling{(Agent Number Scaling)}
\EndFor
\end{algorithmic}
\end{algorithm}
Initialization establishes the foundation for network-centric scaling. The policy network $\pi_{\theta}$ and value network $V_{\phi}$ architectures can be scaled via network size expansion (e.g., increasing layers/width), ensembles (multiple subnetworks), or Mixture-of-Experts (MoE) layers to enhance representational capacity and stability. Concurrently, initializing an agent population $\mathcal{A}$ enables evolutionary scaling strategies, where diversity in policy parameterizations mitigates convergence to suboptimal local minima.  

During environment interaction, data scaling techniques accelerate experience acquisition. Parallelized data collection (e.g., via VectorEnv) maximizes throughput by deploying multiple agents across distributed environments, reducing wall-clock time per episode. Simultaneously, synthetic data generation (e.g., diffusion models or generative adversarial networks) augments the replay buffer $\mathcal{D}$ with high-reward transitions, mitigating exploration bottlenecks in sparse-reward domains. This dual approach --- combining real and generated data --- enhances state-action space coverage while preserving sample validity.  

The model update phase leverages training budget scaling to optimize resource utilization. Replay ratio scaling reuses samples multiple times, amortizing collection costs and improving sample efficiency. Batch size scaling accelerates convergence by stabilizing gradient estimates, while distributed training frameworks parallelize gradient computations across devices for near-linear throughput gains. Further efficiency is achieved via auxiliary tasks (e.g., contrastive or predictive losses), which refine latent representations without additional environment interactions. 

Finally, evolutionary scaling dynamically expands the agent population $\mathcal{A}$. New agents generated through mutation/crossover diversify policy exploration, effectively implementing population-based parallelization.

Although neural scaling laws have been systematically characterized in supervised learning (SL), their extension to DRL remains nascent. This gap arises from fundamental differences in how learning is structured: unlike SL’s static data and well-defined input-output mappings, DRL operates in a dynamic, feedback-driven paradigm where agents interact with environments to generate their own training signals. These differences introduce multi-dimensional challenges that disrupt the straightforward applicability of traditional scaling laws.  

\datascaling{\textbf{Data scaling}} in DRL presents a core challenge distinct from SL. In SL, datasets are typically independent and identically distributed (i.i.d.), enabling predictable scaling through data augmentation or collection. In contrast, DRL agents produce sequential, temporally correlated data through environment interactions, creating a feedback loop where the agent’s current policy dictates the data it observes. This leads to \textit{non-stationarity}: as the policy improves, the data distribution shifts, making it difficult to isolate the impact of increased environment interactions on performance. Compounding this issue is the exploration-exploitation dilemma --- scaling data quantity alone may not improve policies if the agent fails to discover critical states or actions, resulting in suboptimal or degenerate learning trajectories.  

\networkscaling{\textbf{Network scaling}} in DRL further diverges from the trends in SL. While increasing model capacity in SL generally improves performance with sufficient data and compute, DRL faces compounding optimization challenges. Algorithms like policy gradients or Q-learning exhibit inherent instability due to high-variance gradient estimates and non-stationary objectives. Scaling network depth or width can exacerbate these issues, amplifying vanishing gradients or creating saddle points in the loss landscape. Additionally, DRL’s temporal credit assignment problem --- attributing rewards to specific actions over long horizons --- is not inherently resolved by larger networks. Without explicit inductive biases or architectural adaptations, increased model capacity may instead lead to overfitting to recent trajectories or inefficient use of parameters.  

\budgetscaling{\textbf{Training budget scaling}} in DRL introduces additional complexity. The compute scaling of SL is largely governed by epochs and data size, with diminishing returns tied to overfitting. DRL, however, requires balancing multiple interdependent factors: environment interactions (for exploration and data diversity), batch sizes (to stabilize updates amid non-stationary data), replay ratios (to reuse off-policy experiences without overfitting to outdated samples), and auxiliary objectives (such as representation learning or reward prediction). For instance, larger batches may reduce gradient variance but stifle exploration, while high replay ratios risk network's plasticity being damaged. In addition, auxiliary objectives further complicate resource allocation, as they compete with the primary RL objective for optimization bandwidth.These trade-offs create non-monotonic, task-dependent scaling behaviors, precluding universal laws akin to SL.  

Together, these challenges underscore why DRL scaling remains an open problem. The interplay of non-stationary data, unstable optimization, and dynamic compute allocation demands frameworks that account for DRL’s unique feedback-driven nature --- a stark departure from the static, predictable regimes of supervised learning.

\section{Data Scaling}
Data scaling has emerged as a critical paradigm for advancing reinforcement learning (RL), addressing fundamental challenges in sample efficiency, generalization, and computational resource utilization. Its importance manifests through two primary strategies: expanding data collection capacity and synthesizing high-fidelity data. The former enhances training throughput and diversity through distributed sampling, while the latter mitigates data scarcity by generating synthetic transitions that preserve environmental dynamics. 
\subsection{Data Collection}

\begin{figure}[!htbp]
\centering
\includegraphics[scale=0.38]{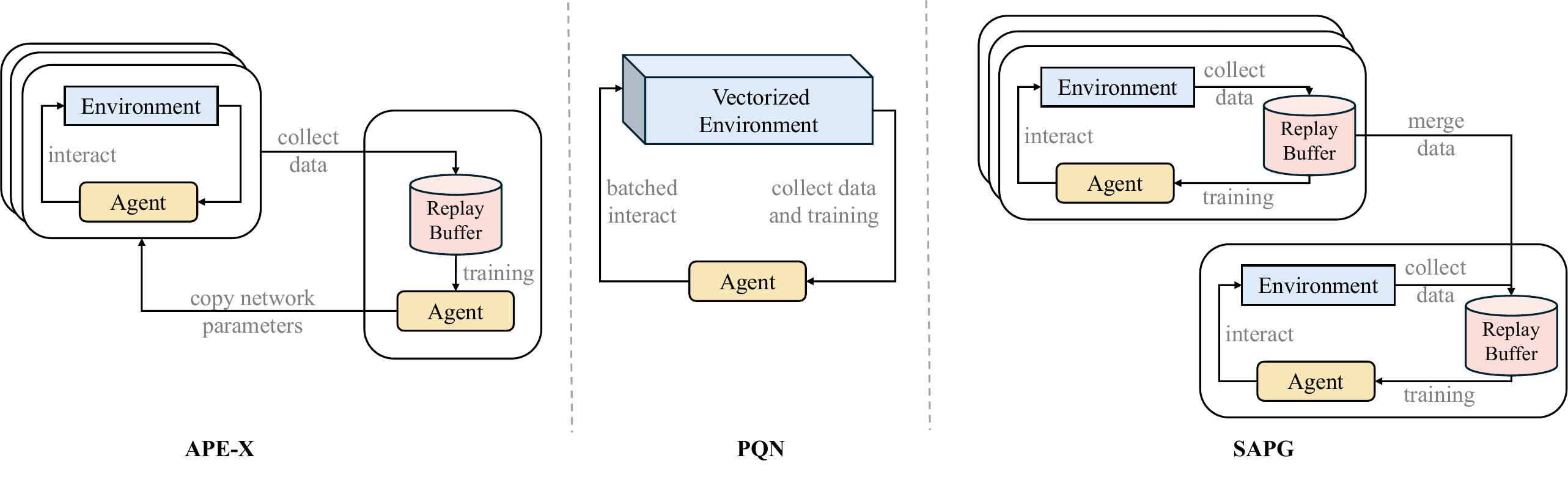}
\vspace{-0.5cm}
\caption{Illustration of different parallel data collecting techniques.}
\label{fig:parallel_data_collecting}
\end{figure}

In online RL, the simplest way in DRL to scale the training data size is to extend the training step \citep{bhattcrossq}. A far more efficient way is to adopt parallel data sampling techniques. By distributing data collection tasks across multiple worker threads or computational nodes, the amount of data collected per unit time is significantly increased. For example, the A3C algorithm employs multiple actors to interact with the environment simultaneously, collecting a large amount of experience data in parallel, thereby reducing the time required for data collection. The parallel data sampling techniques play a crucial role in DRL, offering advantages such as improved training efficiency, stabilized training processes, enhanced exploration capabilities, efficient use of hardware resources, and support for large-scale model training and complex task processing. These advantages help overcome the challenges of low sample efficiency and high computational resource demands, driving the advancement and application of scalable DRL technologies.  An illustration of the difference between different methods is presented in Figure \ref{fig:parallel_data_collecting}.

Several existing works use parallel data sampling. \cite{horgan2018distributed} introduced Ape-X, a fundamental scalable framework for DRL that leverages distributed data generation and prioritized experience replay to efficiently handle large-scale training. In detail, the work decoupled acting (parallel environment interaction with diverse exploration policies) from learning (centralized, prioritized replay of critical experiences). The approach demonstrates that scaling data generation through parallelism significantly enhances learning speed and final policy quality. Similarly,  OpenAI Five \citep{berner2019dota} and AlphaStar \citep{vinyals2019grandmaster} designed distributed infrastructure for the application of RL to complex real-time strategy games with massive parallelization of environment simulations and data collection. \cite{ota2024framework} introduced a framework for effectively training larger networks in DRL by addressing instability and overfitting challenges through  the decoupling of representation learning, the employing of DenseNet  and the leveraging distributed parallel data sampling, i.e., Ape-X to collect diverse on-policy transitions at scale. The results demonstrate scalability across RL algorithms (e.g., SAC, TD3) and environments, highlighting the critical role of distributed data sampling in enabling effective network size scaling for deep RL. Lately, \cite{singla2024sapg} proposed SAPG (Split and Aggregate Policy Gradients), an on-policy reinforcement learning algorithm that addresses scalability limitations in large-scale parallel environments by splitting them into chunks managed by distinct policies and aggregating their data via importance sampling. By employing ensemble-like diversification through latent conditioning and entropy regularization, SAPG enhances exploration and data diversity, outperforming baseline methods like PPO and population-based training in complex robotic manipulation tasks.

In addition to APE-X style and SAPG-style, \cite{gallici2024simplifying} proposed PQN, a simplified deep Q-learning algorithm that leverages parallelized data sampling and regularization techniques to enhance scalability and stability. By replacing traditional components like target networks and replay buffers with vectorized environments for parallel data collection and integrating LayerNorm and $l2$ regularization, PQN achieves competitive performance while reducing computational overhead. Theoretical analysis demonstrates that LayerNorm mitigates instability in off-policy TD learning, and empirical evaluations across Atari\citep{mnih2013playing}, Craftax\citep{matthews2024craftax}, and multi-agent tasks show PQN achieves state-of-the-art sample efficiency with up to 50× faster training compared to conventional methods, highlighting its viability for scalable reinforcement learning. A following-up is the FastTD3 \cite{seo2025fasttd3}, which combines massively parallel simulation, large batches, and distributional critics to solve HumanoidBench tasks in less than three hours on a single GPU.  A recent study \citep{mayor2025impact} further empirically investigates the impact of parallelized data collection—specifically the trade-offs between scaling the number of parallel environments (envs versus rollout length)—on the performance and optimization stability of on-policy DRL algorithms (PPO and PQN). Through extensive experiments, the authors demonstrate that data diversity driven by environmental parallelism is more critical than data volume alone for overcoming plasticity loss and scaling RL agents effectively. 

% In conclusion, 

% \begin{itemize}
%     % \item Crossq: Batch Normalization in Deep Reinforcement Learning for Greater Sample Efficiency and Simplicity
%     \item Don’t Change the Algorithm, Change the Data: Exploratory Data for Offline Reinforcement Learning
%     \item Offline q-learning on diverse multi-task data both scales and generalizes
% \end{itemize}
% RLPD?
% \subsection{Parallel Data Sampling}
% \begin{itemize}
%     \item IMPALA: Scalable Distributed Deep-RL with Importance Weighted Actor-Learner Architectures
%     \item Qt-OPT: Scalable Deep Reinforcement Learning
% for Vision-Based Robotic Manipulation
    % \item A framework for training larger networks for deep
% Reinforcement learning
    % \item Simplifying Deep Temporal Difference Learning
    % \item Distributed prioritized experience replay
    % \item SAPG: Split and Aggregate Policy Gradients
% \end{itemize}

In offline RL, collecting more diverse data from a single task or multi-tasks can also help improve agent's performance.  \cite{yarats2022don}
introduced Exploratory Data for Offline Reinforcement Learning (ExORL), a data-centric framework that emphasizes unsupervised reward-free exploration for collecting diverse single-task or multi-task datasets, which are then relabeled for downstream tasks. By demonstrating that vanilla off-policy RL algorithms outperform specialized offline RL methods when trained on such exploratory data, the work underscores the critical role of data diversity and scale in enabling effective and robust offline policy training.  \cite{kumaroffline} focus on multi tasks. It demonstrated that offline Q-learning, when enhanced with ResNet architectures, distributional cross-entropy losses, and feature normalization, effectively scales with model capacity and data size across diverse multi-task datasets. By training a single policy on 40 Atari games using up to 80 million parameters, the approach achieves human-level performance (over 100\% normalized score) and surpasses supervised methods like decision transformers, particularly in suboptimal data regimes (51\% dataset performance). \cite{park2025horizon} identified the curse of horizon—bias accumulation in TD learning and policy complexity—as the fundamental bottleneck preventing offline RL algorithms from scaling effectively with large datasets (up to 1B transitions) on complex, long-horizon tasks. To overcome this, the authors introduce SHARSA, a hierarchical method that reduces both value and policy horizons via SARSA-based value learning, flow-based behavioral cloning, and rejection sampling for policy extraction. SHARSA achieves state-of-the-art scalability and asymptotic performance.

\subsection{Data Generation}
Recent advances in reinforcement learning have increasingly relied on synthetic data generation to address the fundamental challenge of data scarcity, particularly in offline and sample-constrained online settings. By synthesizing high-fidelity transitions or trajectories that expand dataset coverage while preserving environmental dynamics, generative approaches have emerged as a pivotal strategy for enhancing policy learning without requiring additional real-world interactions.

\cite{wang2022bootstrapped} introduced Bootstrapped Transformer (BooT), a method addressing data scarcity in offline reinforcement learning (RL) by leveraging a Transformer-based sequence model to generate synthetic trajectories, thereby augmenting limited training datasets. By bootstrapping the model with self-generated high-confidence trajectories through autoregressive or teacher-forcing schemes, BooT expands data coverage while maintaining consistency with the underlying Markov decision process, achieving superior performance on D4RL benchmarks \citep{fu2020d4rl}.  Motivated by the work,  \cite{lian2023hipode} later proposed HIPODE, a policy-decoupled data augmentation method for offline reinforcement learning (ORL) that addresses limited dataset coverage by generating high-quality synthetic transitions through negative sampling and value-guided state selection. By employing a CVAE-based state transition model to ensure proximity to the dataset distribution and filtering actions via forward dynamics consistency, HIPODE produces reliable, high-return trajectories that enhance downstream ORL performance without policy dependency. \cite{lu2023synthetic} used a diffusion model-based approach for scaling RL training data by generating synthetic transitions to augment limited real experience, namely Synthetic Experience Replay (SYNTHER). By leveraging synthetic data upscaling, SYNTHER enables effective training of larger policy/value networks in offline RL, improves performance on small datasets, and enhances online RL sample efficiency through higher update-to-data (UTD) ratios without algorithmic modifications. Evaluations across proprioceptive, pixel-based, and latent-space environments demonstrate SYNTHER's ability to match or exceed real-data performance, addressing data scarcity challenges in RL through generative data scaling. Further, \cite{wang2024prioritized} introduced Prioritized Generative Replay (PGR), a method that integrated conditional generative models with relevance-guided synthetic data generation. By training diffusion models to generate transitions prioritized via curiosity-driven or value-based relevance functions, PGR densifies and diversifies training data while mitigating overfitting, enabling higher synthetic-to-real data ratios. The approach demonstrates improved sample efficiency and scalability across state and pixel-based RL tasks over SYNTHER.
 
% TODO: Prioritized generative Replay

% \begin{itemize}
% \item Synthetic Experience Replay
% \item HIPODE: Enhancing Offline Reinforcement Learning with High-Quality Synthetic Data from a Policy-Decoupled Approach
% \end{itemize}

In multiagent RL, there are some works utilizing the intrinsic properties of the multiagent systems, such as permutation invariance, to augment experience data. For example, \citet{ye_experience_2020} generates additional data in MARL by randomly generating a number of permutation matrices to shuffle the unit features in the states. Besides, \citet{gal_esp_2023} take advantage of the global symmetry in the multiagent systems, where rotating the global state results in a permutation of the optimal joint policy, to produce additional samples for effective training.

% \section{Network Scaling}

% Network scaling has become a cornerstone for advancing DRL, enabling agents to tackle increasingly complex tasks by systematically expanding model capacity, enhancing representational power, and optimizing learning dynamics. This paradigm manifests through three synergistic dimensions: architectural innovations that redesign network topologies (e.g., width/depth scaling, transformer integration), algorithmic co-design that introduces tailored training objectives and regularization to stabilize large models, and structural optimization that leverages ensemble strategies and modular components to balance capacity with efficiency. These approaches collectively address the dual challenges of preserving training stability during parameter expansion and unlocking emergent capabilities through scaled architectures—whether through residual connections for value or planning networks, self-attention mechanisms for cross-domain generalization, or ensemble methods for bias-variance control. By transcending conventional architectural constraints, network scaling establishes a scalable pathway to improve sample efficiency, generalization, and task complexity handling in DRL systems. 

\section{Network Scaling}
Network scaling has emerged as a critical paradigm for advancing deep reinforcement learning capabilities through three complementary dimensions: structural expansion of monolithic architectures, ensemble-based parameterization, and evolutionary population dynamics. Structural scaling enhances model capacity by systematically increasing network width or depth while addressing stability challenges through architectural innovations. Ensemble-based approaches aggregate multiple models or modular components to improve robustness, exploration, and uncertainty awareness—transcending single-model limitations through collective intelligence. Evolutionary reinforcement learning (ERL) further extends scaling principles by coordinating populations of agents, synergizing gradient-based optimization with evolutionary strategies to amplify exploration and sample efficiency. These paradigms collectively address fundamental challenges in scalability, generalization, and computational efficiency, each operating at distinct granularities—from neuron-level expansion (network size) and component-wise aggregation (ensembles) to system-level population coordination (ERL).

\subsection{Network Size}
\begin{figure}[!htbp]
\centering
\includegraphics[scale=0.5]{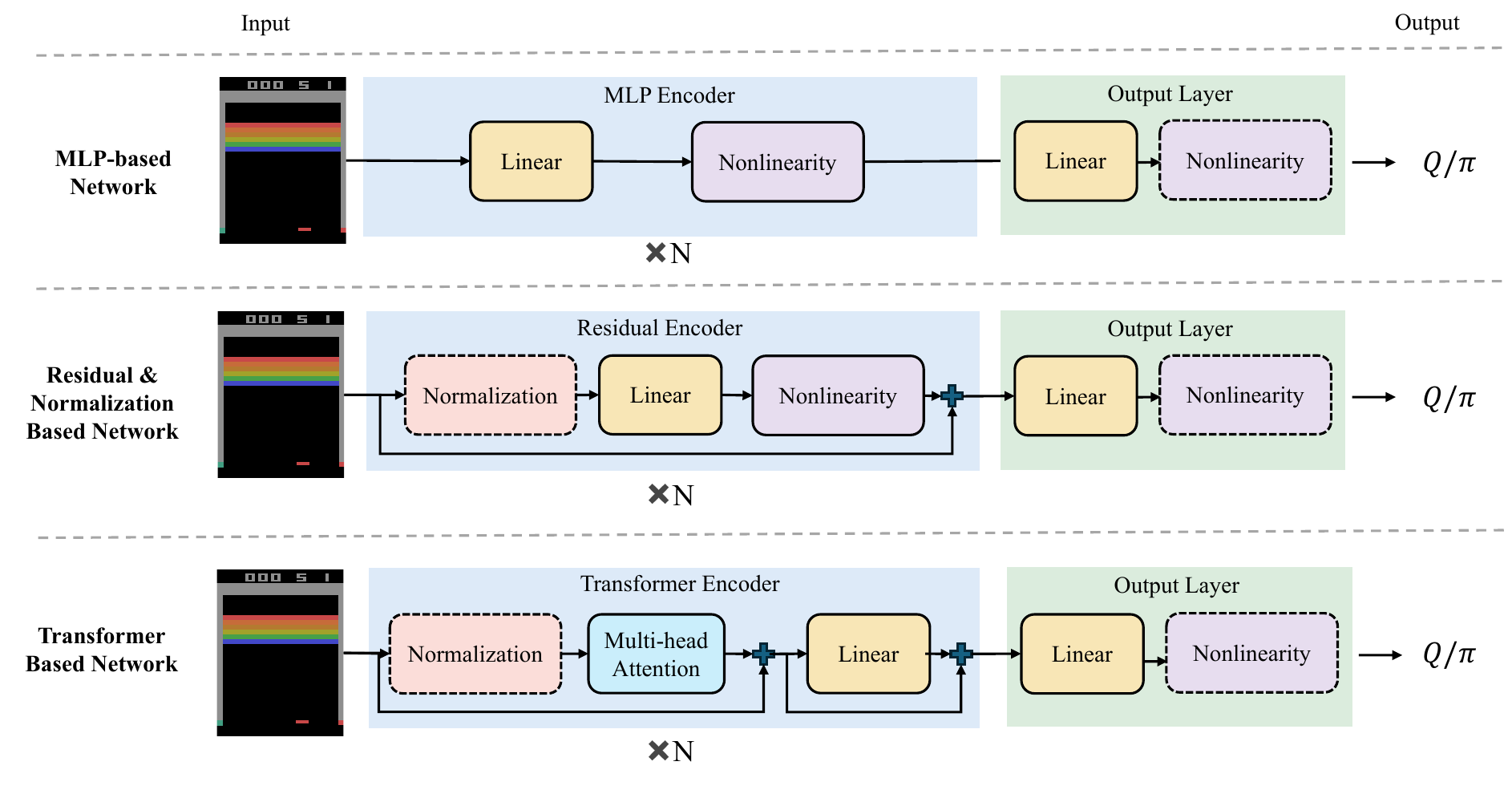}
\caption{Illustration of different network scaling techniques. The dotted lines indicate that the components are optional. The scaling of the network size could be achieved by increasing the dimension of linear layer (i.e., width scaling) or increasing the number of $N$ (i.e., depth scaling).}
\label{fig:network_scaling}
% \vspace{-0.5cm}
\end{figure}

The strategic scaling of neural network architectures has emerged as a pivotal approach to enhance the capacity and efficiency of DRL systems. By systematically expanding model parameters through architectural innovations, researchers have unlocked new performance frontiers while addressing stability challenges inherent to large-scale training. An overall illustration of different network size scaling techniques is presented in Figure \ref{fig:network_scaling}.

\paragraph{Methods using Networks with Residual and Normalization} The methods of first category
focus on scaling MLP-based network's width and depth with architectural enhancements. \cite{ota2020can} explore the impact of network size scaling in DRL by proposing OFENet, an online feature extractor that learns high-dimensional state representations through an auxiliary prediction task. By expanding input dimensionality using a densely connected MLP architecture, the study challenges conventional assumptions that lower-dimensional states are inherently more efficient, demonstrating empirically that higher-dimensional representations improve sample efficiency and policy performance in continuous control tasks. The work highlights the importance of architectural design and feature propagation in scaling neural networks for RL.  \cite{bjorck2021towards} investigate the impact of scaling neural network architectures in DRL, challenging the conventional use of small MLPs in state-of-the-art actor-critic algorithms. The authors identify that training instability caused by exploding gradients—particularly when backpropagating through the critic—hinders the adoption of deeper, modern architectures (e.g., transformers with skip connections and normalization), rather than dataset size or overfitting. By introducing spectral normalization (SN) to enforce Lipschitz continuity in the critic network, they stabilize training and enable effective scaling of model capacity. Their experiments on continuous control tasks demonstrate that scaled architectures with SN outperform traditional RL baselines, highlighting that architectural innovations, alongside algorithmic improvements, are critical for advancing RL performance. \cite{bhattcrossq} propose a method named CrossQ that leverages batch renormalization (BRN) to stabilize training without target networks and employs wider critic layers to improve optimization, achieving state-of-the-art performance in continuous control tasks with a low UTD ratio of 1. This work demonstrates that architectural modifications—such as network width scaling and normalization—can replace complex bias-reduction mechanisms (e.g., high UTD ratios or ensembles).

\cite{naumanbigger} propose BRO algorithm combines regularized scaling of critic networks with techniques like layer normalization and weight decay, alongside optimistic exploration, to achieve high sample efficiency across 40 complex tasks. The results demonstrate that strategic critic network scaling with strong regularization outperforms pure replay ratio scaling, while the combination of both approaches enables unprecedented performance in challenging domains like musculoskeletal control and humanoid locomotion. \cite{ota2024framework} address the challenge of scaling network sizes in DRL by proposing a framework that integrates wider DenseNet architectures, decoupled representation learning via auxiliary prediction tasks, and distributed sampling to mitigate overfitting. Through extensive experiments on continuous control tasks, the authors demonstrate that their approach enables stable training of large networks, achieving significant performance improvements over baseline methods. \cite{lee2024simba} introduces SimBa, a neural architecture designed to scale parameters in DRL by embedding simplicity bias through observation normalization, residual feedforward blocks, and post-layer normalization, which mitigate overfitting in overparameterized networks. By integrating SimBa into diverse RL algorithms (off-policy, on-policy, unsupervised), the approach achieves improved sample efficiency and computational performance, matching or surpassing strong baselines such as BRO across benchmarks like DMC \citep{tassa2018deepmind}, MyoSuite\citep{caggiano2022myosuite}, and HumanoidBench\citep{sferrazza2024humanoidbench}.  Further, \cite{lee2025hyperspherical}  introduce SimbaV2, an upgraded scalable architecture based on Simba through hyperspherical normalization and distributional value estimation with reward scaling. By constraining weight/feature norms and stabilizing gradient dynamics, it effectively scales model capacity (up to 17.8M parameters) and computational budgets (UTD ratios up to 8) while maintaining training stability without requiring periodic reinitialization. The method achieves state-of-the-art performance across 57 continuous control tasks. In addition to the above actor-critic based methods, \cite{schwarzer2023bigger}  introduce BBF, a value-based reinforcement learning agent that achieves super-human performance on the Atari 100K benchmark through strategic neural network scaling and complementary algorithmic innovations. By systematically scaling network width in a ResNet architecture while implementing periodic parameter resets, annealing update horizons, increasing discount factors, and weight decay regularization, BBF demonstrates that network size scaling in deep RL requires careful co-design with sample-efficient training mechanisms. Recently, BRC \cite{nauman2025bigger}, an upgrade of BRO \cite{naumanbigger} is proposed to address optimization challenges in multi-task RL by introducing high-capacity categorical value functions conditioned on learnable task embeddings and trained with cross-entropy regularization. The approach enables robust online temporal-difference learning across 280+ diverse tasks, achieving both model and data scaling.

The above model-free DRL research has predominantly shown that increasing network width generally yields greater performance improvements compared to increasing network depth. In fact, some studies have cautioned that scaling network depth excessively may even have detrimental effects. Against this backdrop, two recent methods have been developed to effectively scale network depth for enhanced planning and self-supervised DRL applications.
\cite{wang2024scaling} introduces Dynamic Transition Value Iteration Networks (DT-VINs), which address network scaling limitations in planning tasks by enhancing the latent MDP's representational capacity through dynamic transition kernels and enabling ultra-deep architectures (up to 5,000 layers) via adaptive highway loss to mitigate vanishing gradients. By substantially increasing both network depth and transition modeling flexibility, DT-VINs achieve state-of-the-art performance in extreme-scale planning scenarios like 100×100 maze navigation and 3D ViZDoom environments requiring 1,800+ planning steps. This work represents a significant advancement in scaling neural network architectures for long-horizon reinforcement learning tasks through systematic improvements to gradient flow and latent state dynamics modeling. \cite{wang20251000} demonstrate that scaling network depth, up to 1024 layers, significantly enhances performance in self-supervised reinforcement learning (RL) for goal-conditioned tasks, achieving 2x–50x improvements in locomotion and manipulation environments. By integrating architectural techniques like residual connections and contrastive RL objectives, deeper networks exhibit emergent capabilities such as complex maze navigation and humanoid acrobatics. The results challenge conventional RL design paradigms by showing depth scaling as a critical factor for unlocking qualitative behavioral shifts and long-horizon reasoning, aligning with scaling trends seen in vision and language models.

\paragraph{Methods using Transformer} Beyond conventional MLP-based scaling, transformer architectures offer unique advantages for cross-task generalization through their inherent scalability and attention mechanisms. \cite{reedgeneralist} introduces Gato, a generalist agent based on a single 1.2B-parameter transformer model, which demonstrates the feasibility of scaling neural networks to handle diverse multimodal tasks—including robotic control, Atari gameplay, and image captioning—within a unified architecture. By training on a vast, heterogeneous dataset and leveraging sequence modeling principles, Gato achieves competitive performance across over 600 tasks, highlighting the potential of network scaling to enable cross-domain generalization. Meanwhile, MGDT \citep{lee2022multi}  investigated the application of large-scale transformer models in multi-task reinforcement learning, demonstrating that a single offline-trained decision transformer achieves near-human performance across 46 Atari games while adhering to scaling laws observed in language and vision domains. The study shows that model performance scales predictably with parameter size (10M to 200M) and enables rapid adaptation to unseen games through fine-tuning, outperforming traditional online RL and temporal difference methods in multi-task generalization. However, Gato and MGDT are essentially transformer-based imitation learning methods. In the field of Transformer-based RL, \cite{parisotto2020stabilizing} proposed the Gated Transformer-XL (GTrXL), a scaled transformer architecture with modified layer normalization ordering and gating mechanisms, to address optimization challenges in DRL. By stabilizing training through identity map reordering and GRU-style gating, GTrXL enables effective scaling to 12-layer networks with 512-step memory, achieving state-of-the-art performance on memory-intensive RL benchmarks (e.g., DMLab-30) while maintaining robustness in reactive tasks. The work demonstrates that architectural innovations can successfully scale self-attention models to deeper configurations in RL settings, outperforming LSTMs and external memory architectures through improved parameter utilization and temporal horizon handling. \cite{springenberg2024offline} demonstrates that offline actor-critic RL can also effectively scale to large transformer-based models, such as the proposed Perceiver-Actor-Critic (PAC) architecture, following scaling laws akin to those in supervised learning. By training on a diverse dataset of 132 continuous control tasks, including real robotics, PAC outperforms behavioral cloning baselines and achieves expert-level performance, particularly when leveraging suboptimal data through RL objectives. The study establishes that model performance improves predictably with increased model size and compute, positioning offline actor-critic methods as a viable pathway for scaling generalist agents in network size scaling paradigms.

\paragraph{Methods using New Training Objectives}
Architectural scaling alone, however, necessitates co-evolution of training methodologies to fully harness expanded network capacities. For example, \cite{farebrother2024stop} proposes replacing regression-based mean squared error (MSE) with categorical cross-entropy classification for training value functions in  DRL, demonstrating significant improvements in scalability and performance across diverse domains. By employing methods like HL-Gauss to convert scalar targets into categorical distributions, the approach mitigates challenges such as noisy targets and non-stationarity, enabling effective scaling with large networks, including Transformers and Mixture-of-Experts. \cite{tang2024improving} investigated the chain effect of value and policy churn in DRL, where non-stationary network updates lead to compounding biases in learning dynamics. The authors propose CHAIN, a regularization method that mitigates churn by minimizing output changes for non-batched states, thereby improving stability and performance across value-based, policy-based, and actor-critic DRL algorithms. Notably, CHAIN enhances network scalability by effectively reducing churn-induced instability when scaling up neural network architectures (e.g., wider/deeper networks).

% \begin{itemize}
%     \item Can Increasing Input Dimensionality Improve Deep Reinforcement Learning
%     \item A framework for training larger networks for deep Reinforcement learning
%     \item Towards Deeper Deep Reinforcement Learning with Spectral Normalization
%     \item Bigger, Regularized, Optimistic: scaling for compute and sample-efficient continuous control
%     \item SimBa: Simplicity Bias for Scaling Up Parameters in Deep Reinforcement Learning
%     \item Hyperspherical Normalization for Scalable Deep Reinforcement Learning
%     \item Crossq: Batch Normalization in Deep Reinforcement Learning for Greater Sample Efficiency and Simplicity
%     \item Bigger, Better, Faster: Human-level Atari with human-level efficiency
%     \item 1000 Layer Networks for Self-Supervised RL: Scaling Depth Can Enable New Goal-Reaching Capabilities
%     \item Scaling Value Iteration Networks to 5000 Layers for Extreme Long-Term Planning
%     \item A Generalist Agent
%     \item Multi-Game Decision Transformers
%     \item Offline actor-critic reinforcement learning scales to large models
%     \item Stabilizing Transformers for Reinforcement Learning
%     \item Stop regressing: Training value functions via classification for scalable deep rl
%     \item Improving Deep Reinforcement Learning by Reducing the Chain Effect of Value and Policy Churn
% \end{itemize}

\subsection{Ensemble and MoE}
\begin{figure}[!htbp]
\centering
\includegraphics[scale=0.5]{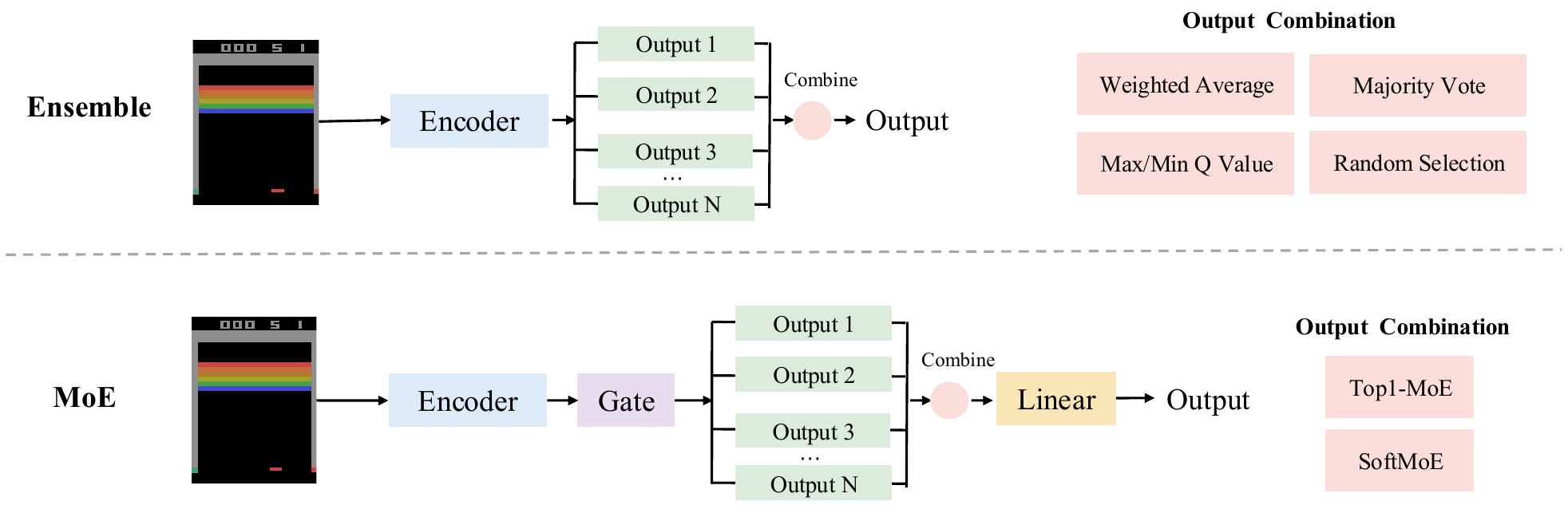}
\caption{Illustration of different network ensemble techniques.In ensemble-based methods, various strategies exist for combining policy and value function outputs: \textit{Weighted Average} assigns learned or heuristic weights to each member’s predictions, prioritizing models with higher confidence or performance. \textit{Majority Vote} selects the action most frequently recommended by individual policies, emphasizing democratic consensus. \textit{Max/Min Q Value} directly adopts the action associated with the highest / lowest Q-value estimate or Q-value with exploration term across ensemble members, prioritizing optimistic or conservative value estimates. \textit{Random Selection} introduces stochasticity by uniformly sampling Q values or actions from the ensemble’s outputs, promoting exploration. \textit{MoE} employs a dynamic gating mechanism to adaptively combine specialized sub-models (experts), enabling context-dependent specialization and enhancing robustness through learned hierarchical integration. \textit{Top1-MoE} employs a simple, efficient routing strategy where each token is assigned to its top-ranked expert, often combined with auxiliary losses to balance expert utilization and prevent specialization collapse. \textit{SoftMoE} replaces traditional discrete token-to-expert assignments with a soft, continuous mechanism, allowing each expert to process learned weighted combinations of all input tokens, improving training stability and computational efficiency.}
\label{fig:network_ensemble_moe_erl}
% \vspace{-0.5cm}
\end{figure}

Ensemble-based network scaling has emerged as a cornerstone of modern reinforcement learning, enabling systematic improvements in robustness, generalization, and algorithmic stability through the principled aggregation of multiple models or components. Modern algorithmic implementations had integrated double critics in basic algorithms such as TD3 \citep{fujimoto2018addressing} and SAC \citep{haarnoja2018soft}. By leveraging diversity across parallel networks, subsampled estimators, or modular architectures, ensemble methods address critical challenges in RL from complementary angles: Q-function ensembles mitigate exploration-exploitation trade-offs and estimation biases through uncertainty-aware value aggregation (\citep{chen2017ucb, lanmaxmin}), while policy ensembles enhance robustness via multi-policy optimization and regularization (\citep{zhang2019ace, schubertpolter}). Beyond functional enhancements, structural innovations like Mixture-of-Experts (MoE) frameworks (\cite{obando2024mixtures})  demonstrate that ensemble-like parameter scaling transcends traditional performance limits, unifying capacity expansion with task-specific specialization. These diverse implementations—spanning online/offline RL, bias-variance control, and architectural redesign—collectively establish ensemble scaling as a versatile paradigm for advancing both theoretical guarantees and practical efficiency in RL systems. An overall illustration of different network size scaling techniques is presented in Figure \ref{fig:network_ensemble_moe_erl}.

\paragraph{Ensemble Q}

Q-function ensemble scaling has emerged as a multifaceted strategy to enhance reinforcement learning robustness, efficiency, and scalability. By aggregating multiple Q-value estimators, these methods address critical challenges such as exploration-exploitation trade-offs (\citep{chen2017ucb}), overestimation bias (\citep{lanmaxmin}), and sample efficiency (\citep{chenrandomized}), while enabling architectural innovations (\citep{obando2024mixtures}). From online RL perspectives, ensembles improve exploration through uncertainty quantification, stabilize value estimation via bias-variance control, and enable high UTD ratios. In offline settings, they mitigate extrapolation errors by penalizing uncertain predictions. Structurally, MoE-based designs demonstrate that ensemble-like parameter scaling can transcend traditional performance limits. Together, these approaches highlight how ensemble scaling—whether through parallel networks, subsampling strategies, or modular architectures—provides a versatile toolkit for advancing both theory and practice in value-based RL.

The integration of ensemble methods for Q-function scaling has evolved through diverse angles in online RL.
\cite{osband2016deep} pioneered ensemble-based exploration via Bootstrapped DQN, which addresses efficient exploration in complex environments by integrating ensemble methods to scale network capacity. By employing multiple parallel Q-network heads with shared convolutional layers and bootstrap sampling, the approach approximates posterior distributions over value functions, enabling temporally extended (deep) exploration akin to Thompson sampling. The ensemble-based design demonstrates statistically and computationally efficient learning, achieving higher cumulative rewards when increasing bootstrap heads. \cite{chen2017ucb} extended the design of Bootstapped DQN. It proposed an exploration strategy for deep Q-learning by leveraging ensembles of Q-functions to estimate upper confidence bounds (UCB), enhancing sample efficiency through uncertainty-aware action selection. By integrating UCB-inspired exploration with Q-ensemble techniques, the method effectively balances exploration and exploitation, achieving superior performance on the Atari benchmark compared to prior approaches like Bootstrapped DQN and count-based exploration. Similarly, SUNRISE \citep{lee2021sunrise} introduces a unified ensemble framework to mitigate instability in off-policy RL, combining ensemble-weighted Bellman backups (re-weighting targets by Q-ensemble uncertainty) and UCB exploration for efficient action selection.

Another line of efforts focused on mitigating estimation biases. Extending Double Q-learning to ensembles, EBQL (Ensemble Bootstrapped Q-Learning) \citep{peer2021ensemble} proposed to mitigate both overestimation and underestimation biases by averaging predictions across multiple Q-networks. \cite{lanmaxmin} addressed the overestimation bias in Q-learning, which arises from using the maximum estimated action value as an approximation for the true maximum value. The authors proposed Maxmin Q-learning, a method that mitigates bias by utilizing the minimum among multiple action-value estimates from an ensemble of N action-value functions, enabling flexible control over estimation bias (from overestimation to underestimation) while reducing variance. Theoretical analysis demonstrates that with an optimal choice of N, the algorithm achieves unbiased estimation with lower variance than standard Q-learning.  TQC \citep{kuznetsov2020controlling} integrates distributional RL, truncation, and ensembling to granularly control overestimation bias in continuous control. By truncating the right tail of a mixture of N distributional critics' quantile estimates and averaging retained atoms, it achieves finer bias regulation than min-ensemble methods. 

Further innovations emphasized diversity and efficiency. 
 \cite{chenrandomized} and \cite{hiraokadropout} balanced bias-variance trade-offs through randomized ensemble subsampling (Randomized Ensembled Double Q-Learning, REDQ) and dropout-based uncertainty injection (DroQ). By employing an ensemble of Q-networks with randomized in-target minimization over subsets of the ensemble, REDQ achieves high UTD ratios, enabling sample efficiency comparable to or surpassing model-based approaches on MuJoCo benchmarks \citep{todorov2012mujoco}. This work demonstrates that ensemble-based scaling effectively balances bias and variance, allowing fewer parameters and faster training than traditional model-based methods. However, REDQ  requires large ensemble numbers, which could bring the computing inefficiency. To this end, \cite{hiraokadropout} proposed DroQ, which enhances computational efficiency while maintaining sample efficiency by employing a small ensemble of Q-functions with dropout connections and layer normalization. By integrating dropout to inject model uncertainty and layer normalization to stabilize training, DroQ achieves comparable performance to REDQ. AdaEQ \citep{wang2021adaptive} adaptively adjusted ensemble size based on time-varying approximation errors, using theoretical bias bounds and error feedback to drive estimation bias near zero. This approach dynamically scales the ensemble to balance computational cost and accuracy, outperforming fixed-size ensembles in MuJoCo tasks by minimizing bias while maintaining efficiency. Similarly, AQE \citep{wu2021aggressive} averages the variable K-lowest Q-values from variable N ensemble members based on REDQ. DNS \citep{sheikh2022dns} reduced ensemble training costs by sampling subsets of networks for backpropagation using determinantal point processes (k-DPP), prioritizing uncorrelated networks to maximize diversity per update. Integrated with REDQ, it cuts computation by 50\% (measured in FLOPS) while matching baseline performance in MuJoCo. \cite{sheikh2022maximizing} further proposed MED-RL, a regularization framework inspired by economic inequality metrics and consensus optimization, to enhance diversity among ensemble members in DRL. By integrating metrics like the Gini coefficient and Atkinson index as regularizers, MED-RL prevents representation collapse in ensemble-based DRL algorithms (e.g., SAC, TD3, MaxminDQN, Bootstrapped DQN, REDQ), ensuring networks maintain distinct learning trajectories. Empirical results across MuJoCo and Atari benchmarks demonstrate significant performance gains (up to 300\%) and improved sample efficiency (75\% faster convergence).

Beyond pure online settings, ensemble scaling principles were adapted to offline RL and offline-to-online challenges.
\cite{an2021uncertainty} introduced an ensemble-based offline reinforcement learning method that leverages clipped Q-learning to address by penalizing uncertain Q-value predictions. It found out that by using minimal Q-value prediction as target Q from the scaled the Q-ensembles, overestimation errors from out-of-distribution (OOD) data could be largely diminished. \cite{lee2022offline} proposed to tackle distribution shift during offline-to-online fine-tuning by integrating a pessimistic Q-ensemble (underestimating OOD actions) and balanced experience replay, which prioritizes near-on-policy transitions using density-ratio estimators. \cite{ball2023efficient} bridged offline-online paradigm by proposing RLPD, an efficient online RL method that leverages offline data through symmetric sampling. It employs critic ensembles with layer normalization to mitigate value over-extrapolation  and improve sample efficiency in complex environments without requiring costly pre-training or explicit constraints. By integrating these minimal yet effective modifications to standard off-policy algorithms, the approach achieves 2.5x improvements in sparse-reward tasks while maintaining computational efficiency.

Finally, structural innovations emerged beyond traditional Q-ensemble. \cite{cobbe2021phasic} introduced Phasic Policy Gradient (PPG), an RL framework that scales the capacity of the network by employing disjoint networks for policy and value functions during the training phases, enabling the optimization of aggressive value functions with greater sample reuse while preserving shared representations through periodic distillation. By alternating between policy optimization (using a clipped surrogate objective) and an auxiliary phase that distills value function features into the policy network via behavioral cloning and auxiliary losses, PPG achieves improved sample efficiency over PPO, particularly in high-dimensional environments like Procgen. \cite{obando2024mixtures} investigated the integration of Mixture of Experts (MoE), specifically Soft MoEs, into value-based deep reinforcement learning (RL) architectures to address the challenge of parameter scalability. By replacing dense layers with Soft MoE modules in networks like DQN and Rainbow, the authors demonstrate that RL agents achieve significant performance improvements on Atari benchmarks as the number of experts increases, contrasting with traditional parameter scaling approaches that degrade performance.

In the domain of multiagent RL, a few preceding works extend ensemble methods from single-agent DRL to reduce Q-value overestimation by discarding large target values in the ensemble. For example, \citet{ackermann_reducing_2019} introduce the TD3 technique to reduce the overestimation bias by using double centralized critics. Besides, \citet{wu_sub-avg_2022} use an ensemble of target multiagent Q-values to derive a lower update target by discarding the larger previously learned action values and averaging the retained ones. More recently, \citet{yang_dual_2025} propose DEMAR to extend REDQ into a dual ensemble Q-learning algorithm to reduce the overestimation in both the individual Q-values and global Q-values with the random minimization operation to stabilize learning.

\paragraph{Ensemble Policy}
Compared to Q ensemble, policy ensemble scaling has been rarely explored. Existing works tried to enhance robustness and generalization by increasing the policy number. \cite{zhang2019ace} proposed ACE (Actor Ensemble Algorithm), a method employs an ensemble of deterministic policy networks (actors) to mitigate local optima when maximizing the critic function, while framing the ensemble within the options framework to formalize each actor as an option with deterministic intra-option policies. ACE further enhances value estimation by integrating a look-ahead tree search using the ensemble's action proposals and a learned value prediction model, demonstrating better performance  than DDPG and its variants in robotic control tasks when using 5 actors. However, it shows that scaling actor number from 5 to 10 leads to performance degradation. Later, SEERL \citep{saphal2021seerl} proposed to generate diverse policy ensembles from a single training instance via directed parameter perturbations, enabling scalable exploration without extra environmental interactions.
Building on multi-policy frameworks, \cite{lyu2022efficient} introduced the Double Actors Regularized Critics (DARC) algorithm, which leverages dual actor networks and regularized critics to address overestimation and underestimation biases in continuous reinforcement learning. By employing double actors for enhanced exploration and value estimation correction, coupled with critic regularization to mitigate uncertainty between dual critics, DARC achieves better sample efficiency and performance on continuous control benchmarks compared to methods like TD3 and SAC. Further expanding applications, \cite{schubertpolter} introduced POLTER (Policy Trajectory Ensemble Regularization), a regularization method for Unsupervised Reinforcement Learning (URL) that leverages an ensemble of policies discovered during pretraining to approximate an optimal prior policy, enhancing sample efficiency for downstream tasks. By minimizing the KL-divergence between the current policy and a mixture ensemble of historical policies, POLTER regularizes pretraining trajectories, reducing deviation from task-agnostic optimal priors and improving state-of-the-art performance on the URL benchmark \citep{laskin2021urlb} for model-free methods. EPPO \citep{yang2022towards} enhanced RL applicability by deploying parameter-shared policy ensembles  trained with diverse exploration strategies (e.g., randomized intrinsic rewards), improving generalization to unseen environments while maintaining parameter efficiency. The ensemble scales policy diversity without proportional computational overhead, leveraging shared feature extractors and decoupled exploration heads to boost sample efficiency in Atari. TEEN \citep{li2023keep} proposed trajectory-optimized ensembles that maximize behavioral divergence via a trajectory discrepancy loss, encouraging policies to cover distinct state-action paths. By scaling ensemble diversity through lightweight parallel policy heads with shared backbone networks, it achieves higher state coverage than SAC in MuJoCo locomotion tasks, accelerating reward discovery without increasing environment interactions.

\begin{table*}[!htbp]
\centering
 \caption{Comparison between ensemble scaling methods.}
	\label{tab:ensemble_scale}

\scalebox{0.8}{
\begin{tabular}{c|ccccc}
\toprule
Method        & \begin{tabular}[c]{@{}c@{}}\textbf{Critic} \\ \textbf{Ensemble}\end{tabular}     & \begin{tabular}[c]{@{}c@{}}\textbf{Policy} \\ \textbf{Ensemble} \end{tabular}            & \begin{tabular}[c]{@{}c@{}}\textbf{Output} \\ \textbf{Combination} \end{tabular}       & \begin{tabular}[c]{@{}c@{}}\textbf{Reported Max} \\ \textbf{Ensemble Number}\end{tabular}      & \begin{tabular}[c]{@{}c@{}}\textbf{Representative} \\ \textbf{Benchmarks}\end{tabular}     \\ \midrule
Bootstrapped DQN \citep{osband2016deep}  &  \faCheck  &      &  Random Selection &   20 &  Atari \\
UCB Explore \citep{chen2017ucb}  &  \faCheck  &       & \begin{tabular}[c]{@{}c@{}} Majority Vote \\ or Max Q Value \end{tabular}  &   10 &  Atari \\
SUNRISE \citep{lee2021sunrise} &  \faCheck &  \faCheck  & \begin{tabular}[c]{@{}c@{}} Mean Q Value \\ + Std Q Value \end{tabular}  &   10 &  Atari,MujoCo, DMC \\
EBQL \citep{peer2021ensemble} &  \faCheck &    & Mean Q Value  &   25 &  Atari \\
Maxmin Q-learning \citep{lanmaxmin} &  \faCheck &    & Min Q Value  &   9 &  Atari \\
TQC \citep{kuznetsov2020controlling}  &  \faCheck &    & Distributional Q  &   2 &  MuJoCo \\
REDQ \citep{chenrandomized} &  \faCheck &    & \begin{tabular}[c]{@{}c@{}} Random Selection \\ + Min Q Value \end{tabular}  &   10 &  MuJoCo \\
DroQ \citep{hiraokadropout} &  \faCheck &    & \begin{tabular}[c]{@{}c@{}} Random Selection \\ + Min Q Value \end{tabular}  &   10 &  MuJoCo \\
AdaEQ \citep{wang2021adaptive} &  \faCheck  &  &   \begin{tabular}[c]{@{}c@{}} Strategically Selection \\ + Min Q Value \end{tabular} &  7 &  MuJoCo \\
AQE \citep{wu2021aggressive} &  \faCheck  &  &   \begin{tabular}[c]{@{}c@{}} Strategically Selection \\ + Min Q Value \end{tabular} &  15 &  MuJoCo, DMC \\
MED-RL \citep{sheikh2022maximizing}  &  \faCheck  &  & Depends on Backbone  &  Depends on Backbone &  MuJoCo, MinAtar \\
MoE\citep{obando2024mixtures}  &  \faCheck  &  & SoftMoE, Top1-MoE  &  8 & Atari \\
EDAC \citep{an2021uncertainty}  &  \faCheck  &  & Min Q Value  &  50 &  D4RL \\
RLPD \citep{ball2023efficient}  &  \faCheck  &  & Min Q Value  &  10 &  D4RL \\
ACE \citep{zhang2019ace}  &   & \faCheck  & Max Q Value  &  10 &  MuJoCo \\
SEERL \citep{saphal2021seerl} &   & \faCheck  & \begin{tabular}[c]{@{}c@{}} Majority Vote \\ or Random Selection \end{tabular} &  9 &  MuJoCo, Atari \\
POLTER \citep{schubertpolter}  &   & \faCheck  & Weighted Average  &  7 &  URL Benchmark \\
EPPO \citep{yang2022towards} &   & \faCheck &  Average  &  8 &  Atari \\
TEEN \citep{li2023keep}  &   & \faCheck &  \begin{tabular}[c]{@{}c@{}} Random Selection \\ + Min Q Value \end{tabular}  &  10 &  MuJoCo \\
DARC \citep{lyu2022efficient}  & \faCheck  & \faCheck  & Max Q Value  &  2 &  MuJoCo \\
\bottomrule
\end{tabular}
}
\end{table*}

\subsection{Agent Number Scaling}

\begin{figure}[!tbp]
\centering
\includegraphics[scale=0.5]{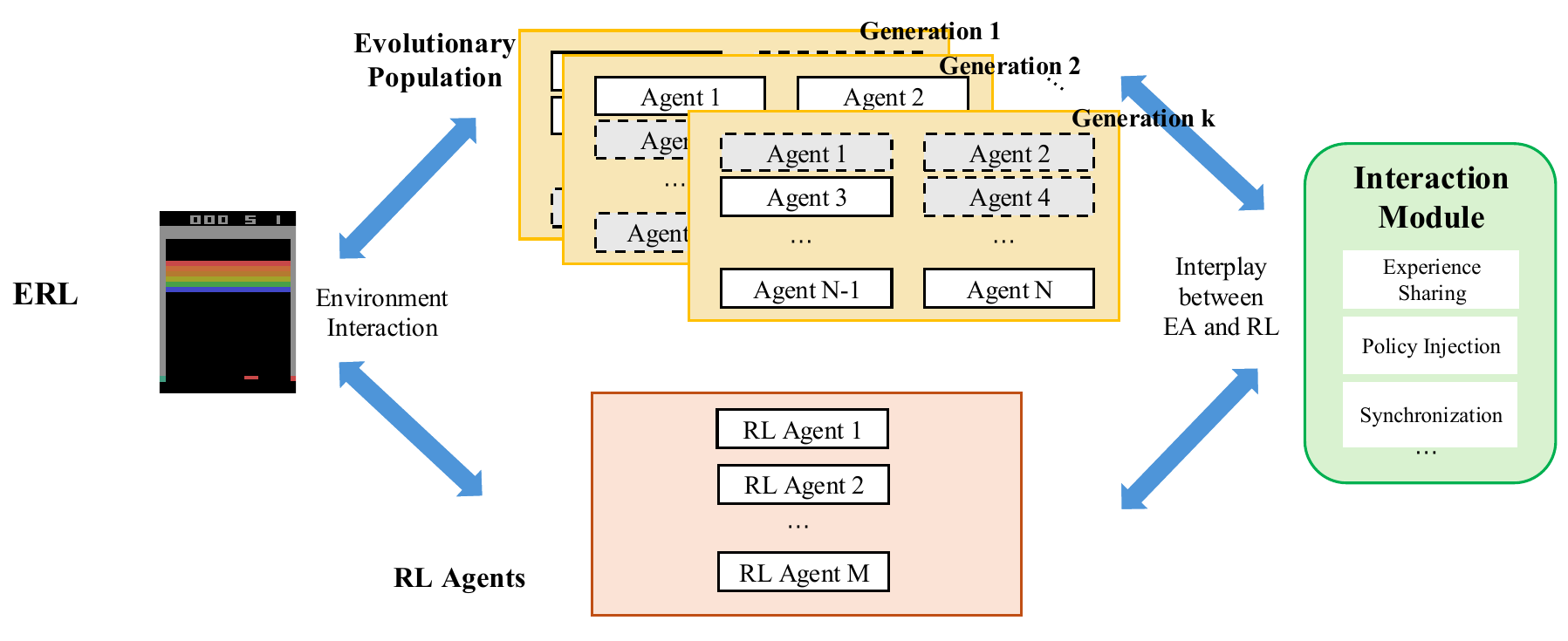}
\vspace{-0.2cm}
\caption{Illustration of Evolutionary Reinforcement Learning (ERL) framework. The framework consists of: an evolutionary population of agents, an RL agent (or multiple RL agents), and an interaction module that controls the interplay between the evolutionary population and the RL agents.
By incorporating the evolutionary population, ERL methods make use of more neural networks than conventional RL methods.}
\label{fig:erl_framework}
\vspace{0.2cm}
\end{figure}%

Except for re-designing network structure to be larger and more expressive monolithic networks or building network ensembles in different ways as surveyed in the previous two subsections, another category of works resort to building a group of deep RL agents. Evolutionary Reinforcement Learning (ERL)~\citep{Li24ERLsurvey} is one of the most representative methods in this literature.
As illustrated in Figure~\ref{fig:erl_framework}, the common framework of ERL mainly consists of three components: (1) an evolutionary population of agents that evolves towards better agent candidates of diversity via evolutionary strategies, (2) an RL agent (or multiple RL agents) that learns according RL algorithms, and (3) an interaction module that controls the interplay between the evolutionary population and the RL agent.
The main idea of ERL is to integrate the advantages of both Evolutionary Algorithm (EA) and RL. EA is known to be inefficient, while RL suffers from difficulties in exploration and gradient optimization; at the same time, EA offers strong exploration and is less sensitive to per-step rewards, and RL offers higher sample efficiency in policy optimization.
The features of both the worlds naturally reveal the potential of combining EA and RL to establish a synergistic learning framework.

By leveraging the evolutionary population of agents, ERL methods make use of more neural networks than conventional RL methods.
Despite the similarity between ERL and ensemble-based RL, we need to note that they operate at different levels: each agent candidate in the population of ERL can be an ensemble-based agent. Therefore, we delineate the separation line between the two categories and introduce the works on network scaling from the angle of ERL in this subsection.
Note that the population size can be much larger in evolutionary learning literature than in ERL literature. Since this work focuses on the scaling in the scope of RL, we exclude the pure evolutionary algorithms.
Besides, population-based training has also been studied and adopted for hyperparameter tuning~\citep{elfwing2018online,franke2021sample}, action selection~\citep{kalashnikov2018scalable,simmons2019q,ma2022evolutionary} in RL. We also exclude these works because they have nothing to do with scaling.

The first work that combines EA and deep RL is ERL~\citep{khadka2018evolution}, which establishes the framework shown in Figure~\ref{fig:erl_framework} and proposes the first instantiation of it at the same time.
In this work, DDPG~\citep{LillicrapHPHETS15ddpg} is adopted for the RL agent and a population of 10 policy networks is used. The policy candidates in the population with high fitness are selected to be the elites and are then probabilistically perturbed through mutation and crossover operations in a Genetic Algorithm (GA) manner to create the next generation of actors progressively.
For the interaction module, the data generated by the evolutionary population is stored in the DDPG agent's replay buffer, which provides diverse samples and thereby enhances sample efficiency.
Conversely, the DDPG agent periodically injects its policy network into the population to be further optimized by GA, which facilitates population evolution with heterogeneous candidates.
In this way, ERL integrates the strengths of both EA and RL. In the experiments, ERL outperforms the basic algorithms DDPG and GA on most OpenAI MuJoCo tasks.
Following ERL, CERL~\citep{khadka2019collaborative} is proposed to further enhance exploration by leveraging multiple RL agents with different discount factors.
The use of multiple discount factors adds myopic and long-term behaviors to the ERL framework.
CERL significantly outperforms ERL in the challenging Humanoid task.
Another follow-up work is GPO~\citep{GangwaniP18GPO}, GPO inherits the ERL framework and replace the parameter-level crossover and mutation operations with learning-based methods.
To be specific, the parameter-level crossover is replaced by network distillation to avoid policy degradation due to the random merging of network parameters.
The parameter-level mutation is replaced by RL for more directed and efficient policy optimization. 
Taking inspiration from GPO, PDERL~\citep{bodnar2020proximal} further improves the crossover and mutation operators in the ERL framework.
PDERL proposes Q-filtered distillation to inherit desirable behaviors from parents to offspring candidates during crossover. PDERL also uses a new mutation operator that adjusts the magnitude of parameter-level mutation according to parameter sensitivity to policy output. Empirically, PDERL demonstrates its superior performance in OpenAI MuJoCo tasks.

In addition to exploring the efficacy of the ERL framework based on off-policy RL agents, GEATL~\citep{GEATL} integrates on-policy RL with EA. Similar to ERL, the on-policy RL agent periodically injects its policy network into the evolutionary population. However, since on-policy RL cannot learn from off-policy samples, EA influences RL in a different manner: the RL agent will synchronize its policy network to the elite policy in the population when the elite policy outperforms the RL policy. GEATL has demonstrated its superiority over ERL in scenarios in Grid World with sparse rewards.
The works mentioned above typically focus on integrating either on-policy RL or off-policy RL with EA. Differently, CSPC~\citep{DBLP:conf/nips/ZhengW0L0Z20} distinguishes itself by simultaneously integrating SAC~\citep{haarnoja2018soft}, PPO~\citep{schulman2017proximal}, and CEM~\citep{StulpS12CEM} to fuse the strengths of three heterogeneous algorithms with distinct update principles.
Specifically, when the SAC policy outperforms PPO or the policies in the population, it supplants those individuals. Similarly, if the PPO policy excels, it takes over the population policies. CSPS outperforms three basic algorithms in most of the MuJoCo tasks.

\citet{Re2} uncovers a primary problem prevalent in the existing ERL methods: the prevalent utilization of isolated policy architectures, where each individual operates within its own private policy network. However, this individualistic learning structure inevitably leads to redundant learning and thus hinders the efficient transfer of valuable insights across the population when using a larger population (i.e., more networks). To address this issue, ERL-Re$^2$ is proposed to decompose the policies into a shared state representation and independent linear policy representations. The policy structure facilitates streamlined knowledge exchange while simultaneously compacting the policy space and enhancing exploration efficiency. Moreover, ERL-Re$^2$ proposes behavioral-level genetic operators based on linear policy representations, coupled with a policy representation-based fitness surrogate to curtail the cost of population evaluation. As a result, ERL-Re$^2$ achieves state-of-the-art performance on MuJoCo tasks.
\citet{Li0TFH24EvoRainbow} provides a systematic comparison among existing ERL methods from the perspectives of the interaction module, the individual architecture, and etc, out of which the best combination of existing design choices is identified and called EvoRainbow. 
By following the same principle of integrating EA and RL, there are also other works that proposes different instantiations of the ERL framework with various methods and designs on the three components of the framework~\citep{kim2020pgps,DBLP:conf/atal/Suri22,DBLP:conf/gecco/ZhengC23}.

\section{Training Budget Scaling}
Training budget scaling constitutes a foundational paradigm for achieving practical and efficient reinforcement learning, addressing computational constraints through three synergistic dimensions: distributed system parallelism, data reuse optimization, and gradient update stability. These strategies collectively maximize the value of limited environmental interactions by reconfiguring how computational resources are allocated spatially (distributed architectures), temporally (experience repetition), and algorithmically (batch processing). Distributed frameworks overcome training bottlenecks through hardware-aware parallelism, while replay ratio scaling transforms temporal data utility via gradient multiplicity. Complementary to these, batch size strategies mediate the stability-efficiency trade-off through variance-controlled optimization. Last but not the least, the auxiliary training methods introduce additional loss to help facilitate the training for higher convergence performance. Together, they enable RL systems to transcend brute-force computational scaling, instead achieving intelligent budget allocation that balances sample efficiency, algorithmic robustness, and hardware utilization across diverse learning scenarios.

\subsection{Distributed Training}
Distributed training has emerged as a pivotal strategy for scaling RL training budgets, addressing computational and data constraints through two synergistic dimensions: system architecture optimization and real-world data scalability. Architectural innovations like decoupled actor-learner frameworks \cite{espeholt2018impala} maximize hardware utilization and throughput via parallelized sampling and learning, while domain-specific implementations \cite{kalashnikov2018scalable} demonstrate how distributed data collection at scale enables robust policy generalization in complex physical systems.

In detail, \cite{espeholt2018impala} introduced IMPALA (Importance Weighted Actor-Learner Architecture), a scalable distributed deep reinforcement learning framework designed to efficiently handle large-scale training across thousands of machines while maintaining high data throughput and resource utilization. By decoupling acting and learning processes and employing the novel V-trace off-policy correction algorithm, IMPALA achieves stable, high-throughput training (250,000 frames/second) and demonstrates superior data efficiency and performance over prior methods like A3C in multi-task settings, including the DMLab-30 and Atari-57 benchmarks.  Expanding beyond simulation, \cite{kalashnikov2018scalable} introduced QT-Opt, a scalable deep reinforcement learning framework for vision-based robotic manipulation, focusing on grasping as a case study. By leveraging a distributed, off-policy Q-learning approach and training on over 580,000 real-world grasp attempts, QT-Opt achieves a 96\% success rate on unseen objects, demonstrating the critical role of large-scale data in learning robust, closed-loop policies. The work underscores how scaling dataset size and parallelized training enables generalization, dynamic behaviors (e.g., regrasping, pre-grasp manipulation), and improved performance over prior methods.

\subsection{Replay Ratio}

The strategic scaling of replay ratio (update-to-data ratio, UTD) has emerged as a pivotal lever for optimizing deep reinforcement learning efficiency, enabling agents to maximize knowledge extraction from limited environmental interactions. By increasing the number of gradient updates per sampled experience, high UTD regimes amplify sample efficiency but introduce critical challenges: compounded estimation biases, catastrophic overfitting, and progressive loss of network plasticity. These challenges necessitate synergistic optimization across architectural design, regularization techniques, and training dynamics—where innovations in bias suppression (e.g., ensemble critics), parameter stabilization (e.g., normalization layers), and plasticity preservation (e.g., periodic resets) collectively unlock the potential of extreme UTD scaling. As modern RL systems increasingly prioritize compute-efficient training, replay ratio scaling operates as both a performance multiplier and a stress test for algorithmic robustness, demanding co-evolution of capacity expansion and optimization stability.

\begin{table*}[!htbp]
\centering
 \caption{Comparison between methods scaling replay ratio.}
	\label{tab:replay_ratio}

\scalebox{0.85}{
\begin{tabular}{c|ccccc}
\toprule
Method        & \begin{tabular}[c]{@{}c@{}}\textbf{Resample} \\ \textbf{Data}\end{tabular}     & \begin{tabular}[c]{@{}c@{}}\textbf{Critic} \\ \textbf{Multiple} \\ \textbf{Update}\end{tabular}            & \begin{tabular}[c]{@{}c@{}}\textbf{Actor} \\ \textbf{Multiple} \\ \textbf{Update}\end{tabular}       & \begin{tabular}[c]{@{}c@{}}\textbf{Reported Max} \\ \textbf{Replay Ratio}\end{tabular}      & \begin{tabular}[c]{@{}c@{}}\textbf{Representative} \\ \textbf{Benchmarks}\end{tabular}     \\ \midrule
REDQ \citep{chenrandomized}  &  \faCheck  &   \faCheck    &   &   20 &  MuJoCo \\
DroQ \citep{hiraokadropout} &  \faCheck  &   \faCheck    &   &   20 & MuJoCo  \\
Primacy Bias \citep{nikishin2022primacy}  &    & \faCheck   &  &  32  & Atari, DMC  \\
SR-SAC/SR-SPR \citep{dsample} &     & \faCheck  &   & 128 & Atari, DMC  \\
BBF \citep{schwarzer2023bigger} &   &  \faCheck   &  & 8  & Atari  \\
CrossQ+WN \citep{palenicek2025scaling} &     & \faCheck  &  &  5 & DMC, MyoSuite  \\
OFN \citep{hussing2024dissecting} &     & \faCheck  &  &  32 & DMC  \\
PLASTIC\citep{lee2023plastic}  &    &  \faCheck &  \faCheck  &  8 & Atari, DMC  \\
AVTD\citep{liefficient2023}   &    &  \faCheck &  \faCheck  &  20 & Atari, MuJoCo  \\
SMR\citep{lyu2024off}  &    &  \faCheck &  \faCheck  &  10 & PyBullet-Gym, DMC  \\
BRO\citep{naumanbigger}  &    &  \faCheck &  \faCheck  &  15 & MetaWorld, DMC, MyoSuite  \\
MAD-TD\citep{voelcker2024mad}  &    &  \faCheck &  \faCheck  &  16 & DMC  \\
Simba\citep{naumanbigger}  &    &  \faCheck &  \faCheck  &  16 & DMC, MyoSuite, HumanoidBench \\
Simbav2\citep{lee2025hyperspherical}  &    &  \faCheck &  \faCheck  &  8 & DMC, MyoSuite, HumanoidBench  \\
\bottomrule
\end{tabular}
}
\end{table*}

Some previously introduced methods with new architecture designs utilize a high replay ratio to further improve the performance. Both REDQ \citep{chenrandomized} and DroQ \citep{hiraokadropout} use an ensemble of Q-functions to suppress the estimation bias brought by the high UTD ratio. Besides, these two methods focus on updating the critic multiple times by sampling different mini-batch each time. Instead, several works enhanced the UTD ratio using fixed batches of experience. Some works focus on updating the critic multiple times using the sampled mini-batch. \cite{nikishin2022primacy} identifies the primacy bias in deep reinforcement learning, i.e., a tendency for agents to overfit early experiences, which is exacerbated by high replay ratios employed to enhance sample efficiency. The authors propose a simple yet effective solution: periodically resetting the last layers of neural networks, which mitigates overfitting while preserving the replay buffer's knowledge, enabling stable training with higher UTD ratios. Empirical results across discrete (Atari) and continuous control (DMC) domains demonstrate that this approach improves performance under high replay ratios, effectively scaling training budgets by allowing more gradient updates per environment interaction without compromising generalization. Further, \cite{dsample} adopted the method in \citep{nikishin2022primacy} to scale the replay ratio. By mitigating neural networks' progressive loss of plasticity, the proposed method enables training with orders-of-magnitude higher replay ratios (up to 128x) than conventional approaches, achieving superior performance on Atari 100k and DeepMind Control Suite benchmarks. BBF \citep{schwarzer2023bigger} combines periodic network resets with architectural innovations (wider ResNet), adaptive n-step horizons, and regularization techniques (weight decay, discount factor annealing) to mitigate overfitting and instability inherent in high-UTD regimes. \cite{palenicek2025scaling}  integrated weight normalization (WN) into the CrossQ \citep{bhattcrossq} framework, which exhibited unstable training dynamics and rapid weight norm growth at elevated UTD regimes. By combining batch normalization with WN to stabilize effective learning rates and prevent loss of plasticity, the proposed CrossQ+WN achieves reliable scaling up to UTD=5 without requiring network resets or drastic interventions. \cite{hussing2024dissecting} proposed output feature normalization (OFN), which projects critic features onto a unit sphere to decouple value scaling from early network layers, enabling high UTD ratio up to 32 based on SAC.

Other works perform multiple policy and value network updates on the same sampled data at each environment step. \cite{lee2023plastic} 
proposed PLASTIC algorithm, which synergistically combines sharpness-aware optimization (SAM), layer normalization, periodic parameter resets, and CReLU activations to preserve both input plasticity (adaptation to shifting data distributions) and label plasticity (adaptation to evolving reward dynamics) under increased update frequencies. Empirical results on Atari-100k and DeepMind Control Suite demonstrate that PLASTIC achieves high sample efficiency and computational Pareto-optimality when scaling replay ratios up to 8×. AVTD (Automatic model selection using Validation TD error) \citep{liefficient2023} identifies high validation temporal-difference (TD) error —reflecting overfitting to the replay buffer—as the primary bottleneck in sample-efficient DRL under high UTD ratios, surpassing challenges like non-stationarity or action distribution shift. To address this, AVTD trains multiple RL agents with diverse regularization strategies (e.g., weight decay, dropout) on a shared replay buffer and dynamically selects the agent with the lowest validation TD error for environment interaction. BRO \citep{naumanbigger} reveals that high UTD ratios synergize with model scaling—larger critics tolerate aggressive gradient reuse (UTD=10–15) without overfitting. Under high UTD, the scaling of the strategic critic network with strong regularization outperforms the scaling of the pure replay ratio. MAD-TD \citep{voelcker2024mad} addresses the instability of high UTD ratio RL by identifying misgeneralization of value functions to unobserved on-policy actions as the primary bottleneck, which exacerbates overestimation and divergence under limited samples. To mitigate this, MAD-TD synthesizes on-policy transitions via a learned world model, augmenting only 5\% of real off-policy replay data with model-generated state-action pairs to regularize value estimation without costly resets or ensembles. This approach stabilizes training at high UTD ratios (up to 16×) and achieves competitive performance with BRO on challenging DDMC tasks. \cite{lyu2024off} proposed Sample Multiple Reuse (SMR). Theoretical analysis and empirical results across 30 continuous control tasks demonstrate that SMR significantly improves sample efficiency over base algorithms by better exploiting collected transitions. SimBa \citep{lee2024simba} demonstrates superior computational efficiency in high UTD regimes, achieving superior performance without complex regularization or planning components, as its architectural bias naturally mitigates overfitting when scaling replay ratios up to 16×. Similarly, thanks to the architecture design, Simbav2 \citep{lee2025hyperspherical} also showed stable performance improvement with the scaling of replay ratios with better convergence results than Simba.

% \begin{itemize}
%     \item Bigger, Better, Faster: Human-level Atari with human-level efficiency
%     \item Bigger, Regularized, Optimistic: scaling for compute and sample-efficient continuous control
%     \item SimBa: Simplicity Bias for Scaling Up Parameters in Deep Reinforcement Learning.
%     \item Hyperspherical Normalization for Scalable Deep Reinforcement Learning
%     \item Scaling Off-Policy Reinforcement Learning  with Batch and Weight Normalization
%     \item Off-Policy RL Algorithms Can be Sample-Efficient for Continuous Control via Sample Multiple Reuse
%     \item Dissecting Deep RL with High Update Ratios: Combatting Value Overestimation and Divergence
%     \item Randomized Ensembled Double Q-Learning: Learning Fast Without a Model
%     \item Dropout Q-Functions for Doubly Efficient Reinforcement Learning
%     \item Sample-Efficient Reinforcement Learning by Breaking the Replay Ratio Barrier
%     \item The Primacy Bias in Deep Reinforcement Learning
%     \item PLASTIC: Improving Input and Label Plasticity for Sample Efficient Reinforcement Learning

Increasing the replay ratio to improve the sample efficiency has also started to receive attention from the MARL community. Recently, \citet{yang_sample-efficient_2024} and \citet{xu_higher_2024} found that a higher replay ratio significantly improves the sample efficiency for MARL algorithms. Notably, \citet{yang_sample-efficient_2024} further investigate the plasticity loss \citep{nikishin2022primacy,lyle_understanding_2023} when training MARL at high replay ratios. They propose MARR to introduce a Shrink \& Perturb strategy \citep{dsample,lyle_understanding_2023} to reset network parameters periodically to maintain the network plasticity and scale a high replay ratio up to 50 under the setting of parallel environments.
    
% \end{itemize}
\subsection{Batch Size}
Batch size scaling serves as a multi-faceted lever to optimize reinforcement learning systems, balancing training stability, hardware utilization, and data-driven optimization through distinct yet complementary strategies. By enlarging batch dimensions, these methods stabilize gradient estimates in continuous control \citep{islam2017reproducibility}, exploit parallel hardware for accelerated training \citep{stooke2018accelerated}, and enhance replay mechanisms via variance-reduced sampling \citep{lahire2022large}. In offline RL, batch scaling emerges as a computationally efficient alternative to ensemble-based uncertainty estimation \citep{nikulin2022q}, while n-step return integration unlocks the latent potential of expanded replay buffers \citep{fedus2020revisiting}. Collectively, these approaches demonstrate that batch size optimization transcends mere computational throughput—it systematically mediates trade-offs between sample efficiency, algorithmic robustness, and hardware-aware scalability across diverse RL paradigms.

\cite{islam2017reproducibility} examined the reproducibility challenges of deep reinforcement learning algorithms, particularly TRPO and DDPG, in continuous control tasks, emphasizing the impact of hyperparameters and training budget allocation. It demonstrates that scaling batch sizes significantly influences performance under fixed computational budgets, with larger batches (e.g., 25,000 for TRPO) improving sample efficiency and stability in environments like Half-Cheetah, while smaller batches lead to suboptimal convergence. Building on this, \citet{stooke2018accelerated} introduced a parallelized framework to efficiently utilize modern hardware (CPUs/GPUs) through synchronized sampling and large batch training. By adapting algorithms like A2C, PPO, DQN, and Rainbow to leverage massively parallel environments and multi-GPU optimization, the authors demonstrate that training with significantly larger batch sizes (e.g., up to 2048) does not degrade sample complexity or final performance when paired with appropriate learning rate scaling and optimization techniques (e.g., Adam). Their approach reduces wall-clock training times drastically (e.g., from days to hours on Atari benchmarks), enabling rapid experimentation while maintaining sample efficiency.

The interplay between batch scaling and data storage and sampling mechanisms further refines its benefits. \cite{fedus2020revisiting} investigated the interplay between experience replay mechanisms and algorithmic components in deep reinforcement learning, focusing on replay capacity and replay ratio. The study reveals that increasing replay capacity significantly enhances performance in algorithms utilizing n-step returns (e.g., Rainbow), while conventional methods like DQN show no improvement, highlighting the critical role of n-step returns in leveraging larger replay buffers despite their theoretical off-policy limitations.  \cite{lahire2022large} addresses the challenge of non-uniform sampling in experience replay for DRL by framing it as an importance sampling problem to minimize gradient variance. The authors propose Large Batch Experience Replay (LaBER), which approximates the theoretically optimal (but intractable) gradient-norm-based sampling distribution by pre-sampling a large batch, computing surrogate priorities, and downsampling to reduce variance. Empirical results across Atari games and continuous control tasks demonstrate that LaBER improves convergence speed and performance over Prioritized Experience Replay (PER) and uniform sampling, while maintaining computational efficiency through its scalable large-batch approximation. 

Finally, \cite{nikulin2022q} extended above principles to offline RL by leveraging large-batch optimization to accelerate Q-ensemble methods, addressing the computational inefficiency of training large ensembles. By scaling mini-batch sizes with square-root learning rate adjustments, the method reduces training time by 3-4x while maintaining state-of-the-art performance on D4RL benchmarks. The results demonstrate that batch scaling effectively replaces ensemble scaling, offering a practical approach to training budget optimization in offline RL without sacrificing algorithmic conservatism.

% \begin{itemize}
%     \item Q-Ensemble for Offline RL: Don't Scale the Ensemble, Scale the Batch Size
%     \item Accelerated methods for deep reinforcement learning
%     \item Reproducibility of benchmarked deep reinforcement learning tasks for continuous control
%     \item Large batch experience replay
%     \item Revisiting fundamentals of experience replay
% \end{itemize}
% \begin{itemize}
    
% \end{itemize}

\subsection{Auxiliary Training}

Incorporating auxiliary training alongside the conventional RL training process is another way to improve deep RL by leveraging extra training budgets.
In essence, this is because vanilla RL is inefficient due to the \textit{tabula rasa} nature, i.e., a conventional RL agent learns from scratch with no prior knowledge at the beginning of learning.
One representative problem is visual RL, where the observation space is pixel- or image-based. This requires the RL agent to extract the decision-related information from visual observations and then learn its policy.
For example, different from human beings who naturally understand the semantics, the physics, the game content and logic in the Atari Pong game, an RL agent does not possess the information in its network and needs to learn solely from reward signals, thus being inefficient apparently.
To this end, various methods are proposed to incorporate auxiliary training in conventional RL by injecting additional signals, aiming to learn useful representations and facilitate policy learning. 

One major stream of work on incorporating auxiliary training focuses on unsupervised representation learning in RL.
\citet{Yarats0KAPF21SACAE} proposes SAC+AE for visual robot locomotion tasks in DeepMind Control Suite (DMC)~\citep{Tassa2018DMC}. Rather than learning from visual observations directly, SAC+AE includes an auxiliary reconstruction task of visual observation by building a branch network based on VAE~\citep{KingmaW13VAE,HigginsMPBGBML17betaVAE}. The additional training on the reconstruction objective helps to extract a compact representation of visual observation, which largely improves the learning efficiency of SAC algorithm.
Beyond unsupervised reconstruction, \cite{LaskinSA20CURL} incorporates contrastive learning in SAC and proposes CURL. CURL generates positive and negative samples of the visual observation (i.e., image) with data augmentation by applying random shift to the observation images. An auxiliary objective is then constructed according to InfoNCE loss~\citep{InfoNCE} and MoCo architecture~\citep{He0WXG20MOCO}.
With the auxiliary contrastive learning, the RL agent learns a latent representation space where the representations of positive samples lie close to each other and the representations of negative samples lie far apart, which facilitates the generalization of similar states and leads to superior performance in DMC tasks. 
Similarly, more auxiliary training methods are further proposed by incorporating general unsupervised representation ideas in RL, e.g., contrastive learning~\citep{AnandROBCH19STDIM,ZhuXWDZQLL23MCURL}, data augmentation~\citep{YaratsKF21DrQ,LaskinLSPAS20RAD,YaratsFLP22DrQv2}, etc.

Different from incorporating auxiliary training based on general unsupervised representation learning, another notable stream of works designs more RL-oriented auxiliary tasks.
\citet{JaderbergMCSLSK17UNREAL} makes very first attempts in this direction and proposes a method called UNREAL. UNREAL incorporates several auxiliary training objectives alongside the conventional on-policy learning objectives in A3C~\citep{MnihBMGLHSK16A3C}, including reward prediction, pixel change maximization and additional off-policy value network training.
These auxiliary training objectives provide additional gradients to the CNN and LSTM layers of the A3C agent.
UNREAL significantly improves both the convergence performance and sample efficiency of A3C in Labyrinth, a first-person 3D game with visual observations and Atari games as well.
\cite{Amy21DBC} revisits the bisimulation metric~\citep{LarsenS89Bisimulation} in the literature of state abstraction and proposes DBC in the context of deep RL.
By learning a dynamics transition model and a reward model, the bisimulation metric is computed approximately. The DBC auxiliary objective is then established for an SAC agent to learn a latent representation space for visual observation that obeys the bisimulation metric.
The auxiliary training of DBC effectively makes the RL agent learn an invariant representation that extract the essential features related to policy learning and exclude distracting factors at the same time. Therefore, DBC improves the learning performance of SAC in DMC tasks, especially when the visual observation contains noisy and unrelated factors.
Concurrently, \cite{SchwarzerAGHCB21SPR} proposes SPR, which utilizes the auxiliary objective of minimizing the prediction error between the true future states and the predicted future states in the representation space, with the help of a multi-step dynamics transition model.
With the auxiliary training of SPR, the RL agent learns the representation of visual observation that is predictive of the environmental dynamics in future steps that could be made by the agent itself. The representation thus favors policy learning and value network training in the conventional RL training process.
SPR significantly improves the learning performance and sample efficiency in Atari tasks.
PlayVirtual~\citep{YuLZFZC21PlayVirtual} is a follow-up work of SPR that extends the auxiliary training by introducing virtual reverse dynamic transition trajectories and imposing cycle consistency on the representation to learn, which further improves learning performance.
Beyond the scope of learning representations of state or observation, auxiliary training is also incorporated for the representations of action and policy, which significantly improves policy performance in large complex action spaces~\citep{Chandak19learningaction,LiTZHLWMW22hyar} and accelerates the iterative policy learning by generalizing value functions among policy space~\citep{Tang2022pevfa,Zhang2022ptheory}.

\section{Discussion}  
\subsection{Insights from Scaling Paradigms}  
The systematic scaling of data, networks, and training budgets has fundamentally reshaped the landscape of DRL for decison making, offering distinct yet complementary pathways to address core challenges in sample efficiency, generalization, and computational demands. Our analysis reveals critical trade-offs and synergies across these dimensions, providing actionable insights for practitioners and researchers.  

\paragraph{Data Scaling: Quantity vs. Fidelity}  
Data scaling strategies bifurcate into parallel collection and synthetic generation, each with unique strengths. Distributed sampling (e.g., Ape-X \citep{horgan2018distributed}) excels in online settings by maximizing environmental interaction throughput which is critical for exploration-heavy tasks requiring diverse real-world interactions. However, its hardware dependency and synchronization overhead limit accessibility for resource-constrained scenarios. In contrast, synthetic data generation (e.g., SYNTHER \citep{lu2023synthetic}) circumvents environmental bottlenecks but risks distributional shift if generative models inadequately capture transition dynamics. Hybrid approaches like prioritized generative replay \citep{wang2024prioritized} demonstrate that coupling real and synthetic data with relevance filtering can mitigate this risk. Practitioners should prioritize parallel collection when environment interaction costs are low, reserving synthetic augmentation for domains with expensive interactions (e.g., robotics) or offline RL with fixed datasets.  

\paragraph{Network Scaling: Capacity vs. Stability}  
Architectural scaling reveals a fundamental tension between representational power and training stability. While wider networks consistently improve performance across model-free algorithms \citep{bhattcrossq}, excessive depth introduces vanishing gradients unless mitigated through highway connections \citep{wang2024scaling} or contrastive objectives \citep{wang20251000}. Transformer-based architectures \citep{reedgeneralist} unlock cross-task generalization but demand orders-of-magnitude more data than MLPs, making them ill-suited for sample-constrained settings. Ensemble and ERL methods provide a middle ground: they enhance exploration and bias reduction with modest parameter increases. For practitioners, width scaling with normalization \citep{lee2024simba} offers the most reliable baseline, with depth or attention mechanisms reserved for long-horizon planning or multi-task generalization.  

\paragraph{Training Budget Scaling: Efficiency vs. Plasticity}  
High replay ratios \citep{naumanbigger} and large batch training \citep{lahire2022large} maximize hardware utilization but induce primacy bias \citep{nikishin2022primacy} --- where agents overfit early experiences. Architectural co-designs like hyperspherical normalization \citep{lee2025hyperspherical} and plasticity-preserving regularization \citep{lee2023plastic,tang2025mitigating} mitigate these risks, enabling stable training with high UTD ratios. Distributed training accelerates training via parallel sampling and learning, boosting hardware utilization.  Auxiliary training injects additional signals, such as reward prediction or contrastive learning, to enhance learning efficiency and plasticity. Together, these approaches ensure that reinforcement learning systems can learn efficiently, stably, and flexibly within limited resource constraints. 

\subsection{Scaling RL in Large Language Models} 
Beyond the scope of typical decision-making problems focused on by canonical RL research, scaling RL is playing a critical role in the post-training of Large Language Models (LLMs) to strengthen reasoning, alignment and generalization abilities from multiple scaling perspectives at training time and test time, as evidenced by recent research breakthroughs.

% \paragraph{Model, Data and Training Budget Scaling at Training Time} 
\paragraph{Model Scaling at Training Time} 
 The pioneer explorations identify several RL scaling dimensions during training. Model size is a primary driver, where larger parameter counts (e.g., OpenAI's o3 model~\citep{openaio3}, Kimi's 1T-param model~\citep{kimik2}, DeepSeek's 671B-param model \citep{deepseekai2025deepseekv3technicalreport} and Qwen3's 235B-param model~\citep{qwen3}) significantly boost reasoning and alignment capabilities at the demand substantially greater computational resources. To optimize efficiency, reward models are often scaled down relative to the main model (e.g., Kimi's 6B reward models \citep{team2025kimi}), maintaining alignment quality while reducing training overhead. 
 Perhaps somewhat special to LLMs, context window length and maximum response rollout length are two other important dimensions for scaling. The former determines the range of historical context that can be conditioned on for text generation, while the later determines the size of the solution space, i.e., longer rollout length allows possible responses for more complex questions. Especially for RL fine-tuning of LLMs, rollout length is extended to enable complex multi-step reasoning, with sequences scaled to 128K tokens in models like Kimi to handle intricate tasks such as mathematical proofs and code generation.
 In practice, this is often achieved by multi-stage training~\citep{qwen3,deepscaler2025,Polaris2025}, where at the early stage a relatively shorter rollout length is used to balance the training budget, and the rollout length is extended in the later stages of training for stronger reasoning behaviors.

\paragraph{Data Scaling at Training Time} 
 In addition to model size, data is another critical factor to the efficacy of RL fine-tuning for LLMs.
 DeepScaleR~\citep{deepscaler2025} presents a high-quality dataset for math reasoning, and Polaris~\citep{qwen3,deepscaler2025,Polaris2025} further proposes filtering data dynamically according to the difficulty of the corresponding reasoning question to solve.
 Recent work~\citep{liu2025scalingrlunlockingdiverse} shows that the quantity and quality of training data derived from extensive human judgments directly enhances reward model accuracy, with over 100K high-quality samples used to refine feedback mechanisms, consequently showing that further prolonged RL fine-tuning improves the reasoning ability of LLM over a range of reasoning tasks like math, coding, and logic puzzle. 
 From another angle, off-policy RL fine-tuning is being pursued by recent studies for more thorough utilization of data through off-policy training.
 RePO~\citep{li2025repo} is proposed based on GRPO~\citep{Shao24GRPO} to replay both historical off-policy data and on-policy data together with different off-policy data replay strategies.
 Differently, ReMix~\cite{liang25squeeze} proposes mix-policy proximal gradient method along with an increased UTD to leverage the early-stage efficiency boost of off-policy learning, while adopting a policy reincarnation strategy for later-stage asymptotic performance improvement.
 Seeking useful off-policy data outside the learning model itself, LUFFY~\citep{yan2025learning} uses off-policy samples from superior models (e.g., DeepSeek-R1) and takes them as demonstrations to shape the LLM policy.

 \paragraph{Training Budget Scaling at Training Time} 
 As to the dimension of training budget, the performance of LLMs is continually improved during the process of RL fine-tuning, which has been widely observed~\citep{deepseekai2025deepseekv3technicalreport,qwen3}. 
 In addition, batch size is increased (e.g., 512 with 64 minibatches in Kimi \citep{team2025kimi}) to stabilize training gradients and accelerate convergence, leveraging distributed infrastructure for efficient parallel processing. 
 The gradient update budget dictates the number of training iterations (e.g., Kimi's 256K episodes \citep{team2025kimi}), balancing exploration and exploitation while mitigating overoptimization through techniques like KL-penalty scheduling. 
 
 Collectively, these dimensions --- model architecture, data engineering, and computational allocation --- constitute a synergistic framework for scaling RL in LLMs, enabling performance gains while managing resource tradeoffs.

\paragraph{Inference Budget Scaling at Test Time}
Unique to LLMs, test-time scaling indicates that the model is more likely to solve the problem with more inference budget at test time.
This is essentially due to the reflection and in-context generation abilities of LLMs.
The evidence for this kind of inference budget scaling characteristic has been presented by notable models like o1~\citep{openaio1} and Qwen3~\citep{qwen3}.
\citet{Muennighoff25s1testtime} shows that manually interrupting the reasoning process of LLMs and inserting explicit reflection tokens also elicit better reasoning results as the scaling of inference budget.
This allows more sophisticated strategies for leveraging the test-time scaling characteristic of LLMs to be proposed.

\paragraph{Agent Number Scaling at Test Time} One promising direction is that LLM can scale the number of agents in the test-time scaling to tackle complex tasks, which is similar to scaling single-agent reinforcement learning to multiagent reinforcement learning. For example, \citet{lifshitz_2025_multiagentverification} introduce Multi-Agent Verification (MAV) as a test-time compute paradigm that combines multiple verifiers for improving language model performance at test-time. Another closely related topic that receives much attention nowadays is the LLM-based multiagent debating \citep{du_2024_mad,chan_2024_chateval,liang_2024_encouraging}, where multiple LLM-based agents are organized to be engaged in a debating process to verify and refine the LLM outputs from different angles of agents. This kind of multiagent debating approach shows potential to enhance mathematical and strategic reasoning across a number of tasks. On the other hand, there are also some works utilizing a collection of agents to simulate human social activities such as simulating believable human daily-life behavior \citep{park_2023_generative} and the entire process of hospital \citep{li_2025_agenthospital} involving patients, nurses, and doctors driven by LLMs.

\subsection{Open Problems and Future Directions}  
Despite remarkable progress, fundamental challenges persist in scaling paradigms, necessitating coordinated advances across algorithmic, theoretical, and infrastructural fronts.  
 
\paragraph{Unclear Interdependencies between Scaling Dimensions}  
Most existing works treat data, network, and training budget scaling as independent axes, but their interdependencies are poorly understood. For instance, synthetic data generation’s effectiveness depends on policy network capacity \citep{wang2022bootstrapped}, while optimal UTD ratios vary nonlinearly with critic width \citep{naumanbigger}. Adaptive frameworks that jointly optimize these dimensions --- perhaps through meta-learning or differentiable architecture search --- could help understand the relationships of different scaling dimensions.  

\paragraph{The Scalability-Efficiency Paradox}  
While network scaling laws \citep{springenberg2024offline} mirror those in supervised learning, DRL uniquely suffers from non-stationary objectives and delayed credit assignment. The emergent capabilities observed in 1000-layer networks \citep{wang20251000} suggest untapped potential, but current architectures still face the challenges in scaling depth under the reward signal. In addition, it lacks theoretical justification for their depth/width configurations. A unified theory of capacity scaling in RL --- connecting representational power, optimization landscape geometry, and exploration-exploitation trade-offs --- remains an open challenge.  

\paragraph{Cross-Modal Scaling Disparities}  
Vision-based RL lags behind state-based methods in scaling benefits, as pixel inputs exacerbate overfitting when increasing network capacity \citep{schwarzer2023bigger}. While latent-space approaches \citep{lu2023synthetic} partially alleviate this, fundamental limitations persist in scaling spatial-temporal representations. Integrating recent advances in video diffusion models with RL-specific compression could unlock new scaling frontiers for embodied AI.  

\paragraph{Benchmarking and Evaluation Gaps}  
Current scaling studies focus on narrow task distributions (e.g., DMC \citep{lee2024simba}), obscuring generalization to open-world complexity. Standardized benchmarks assessing 1) cross-domain transfer, 2) compositional reasoning, and 3) robustness to distribution shifts are urgently needed. Furthermore, the community lacks consistent metrics for scaling efficiency --- proposals should integrate sample complexity, computational cost, and energy consumption into multi-objective evaluations.  

In conclusion, the scaling paradigm has propelled DRL for decision making into new performance regimes, but its full potential remains constrained by fragmented optimization strategies and incomplete theoretical foundations. Addressing these open problems demands holistic approaches that unify data, architecture, and computation through the lens of dynamical system theory and resource-aware learning.

\section{Conclusion}  
The systematic application of scaling laws has emerged as a transformative force in deep reinforcement learning (DRL), offering principled pathways to overcome longstanding challenges in sample efficiency, generalization, and computational demands. This review synthesizes progress across three foundational dimensions --- data, network architectures, and training budgets --- demonstrating how strategic scaling redefines the capabilities of DRL systems. By analyzing emerging methodologies, trade-offs, and unresolved challenges, we establish a framework for advancing scalable, robust, and generalizable RL agents.  

First, data scaling strategies reveal critical insights into balancing quantity and fidelity. Distributed sampling frameworks like Ape-X \citep{horgan2018distributed} and synthetic generation techniques such as SYNTHER \citep{lu2023synthetic} exemplify how parallelized collection and generative augmentation expand data diversity while addressing environmental interaction costs. However, the interplay between dataset scale, exploration dynamics, and distributional alignment underscores the need for hybrid approaches that integrate real and synthetic experiences. Second, network scaling demonstrates that capacity expansion --- through wider architectures \citep{lee2024simba}, transformer-based models \citep{reedgeneralist}, and ensemble methods \citep{chenrandomized} --- enhances representational power but necessitates innovations in normalization and regularization to preserve stability. Third, training budget optimization via high replay ratios \citep{naumanbigger}, large batch strategies \citep{lahire2022large}, and distributed systems \citep{espeholt2018impala} highlights the delicate balance between computational efficiency and algorithmic plasticity, where architectural co-design mitigates overfitting and primacy bias.

Future progress hinges on addressing these open problems through interdisciplinary innovations. Theoretical frameworks must formalize the relationship between scaling dimensions and emergent capabilities, while benchmarks should evaluate cross-domain transfer and compositional reasoning. Advances in generative modeling, meta-optimization, and hardware-aware parallelism will further enable adaptive scaling tailored to task complexity and resource constraints. As DRL transitions toward embodied AI and multi-agent systems, intentional integration of scaling principles will be pivotal in realizing agents that generalize across environments, adapt to dynamic constraints, and learn efficiently from limited interactions. By bridging algorithmic advances with systemic scalability, the next generation of DRL systems promises to unlock unprecedented capabilities in autonomous decision-making.

\bibliographystyle{plainnat}
\bibliography{main_arxiv}
\end{document}